\documentclass[journal]{IEEEtran}
\usepackage{graphicx}
\usepackage{amsmath}
\usepackage{pifont}
\usepackage{epsfig}
\usepackage{amsfonts}
\usepackage{longtable}
\begin{document}
%
\title{Sparse Representation of a Blur Kernel for Blind Image Restoration}
%
%
\author{Chia-Chen~Lee,
        and~Wen-Liang~Hwang\\
        Institute of Information Science, Academia Sinica, Taiwan\\}

\maketitle

\begin{abstract}

Blind image restoration is a non-convex problem which involves restoration of images from an unknown blur kernel. The factors affecting the performance of this restoration are how much prior information about  an image and a blur kernel are provided and what algorithm is used to perform the restoration task.
Prior information on images is often employed to restore the sharpness of the edges of an image.  By contrast, no consensus is still present regarding what prior information to use in restoring from a blur kernel due to complex image blurring processes. In this paper, we propose modelling of a blur kernel as a sparse linear combinations of basic 2-D patterns.  Our approach has a competitive edge over the existing blur kernel modelling methods because our method has the flexibility to customize the dictionary design, which makes it well-adaptive to a variety of applications.
As a demonstration, we construct a dictionary formed by basic patterns derived from the Kronecker product of Gaussian sequences.  We also compare our results with those derived by other state-of-the-art methods, in terms of peak signal to noise ratio (PSNR).

\end{abstract}


%
\IEEEpeerreviewmaketitle

\section{Introduction}
%
%
%
%
\PARstart{I}{mage} blurring is a common problem in digital imaging  generally when pictures are taken with wrong focal length, camera shake, object motion, or shallow depth of field, to name a few \cite{CZF10,Jia07}. In this regards, blind image restoration problem involves restoration of image $X$  from a noisy observation $Y$, which contains less information provided by blur kernel $H$:
\begin{equation}
Y = H *_2 X + N, \label{blurreq}
\end{equation}
where $N$ is noise, and $*_2$ indicates the 2-D convolution operation. However, 
the blind restoration method is difficult as it is impossible to restore $X$ without a simultaneous restoration of $H$. Since the restorations of $H$ and $X$ are not jointly convex, 
the solution of blind restoration is highly dependent on the initial guess of $H$ or $X$. This dependence can be lessened, but not completely removed, even through the imposition of convex regularizations on $X$ and $H$.

Several attempts have been made to regularize image $X$, and all of them aimed to recover local high frequency components (edges) of $X$ from image $Y$. 
The widely adopted total variation approach, which implicitly models $X$ as a piecewise smooth function,  intends to derive an image having a small total variation. Sparse representation models $X$ as an element in a subspace which is spanned by a few unknown atoms in a dictionary. By contrast, there is no widely accepted regularization on the blur kernel $H$. Regularizations on $H$ vary widely from a parametric form (a Gaussian function with unknown standard deviation for handling out-of-focus blurring) to $l_1$-norm on the support of $H$ (for handling motion blurring). In this paper, we propose a novel approach by imposing on $H$ a union of  subspace models. In this approach, we assume that  $H$ can be sparsely represented over a dictionary of atoms. 
We believe that this approach can set a connection between the various forms of regularizations on $H$ through the dictionary design.  This approach will not only incorporate the parametric approach of imposition on $H$ (through assumption of a dictionary of atoms, with each atom derived by certain parameter values),
but it will also allow $H$ freedom to adapt to various applications, because the dictionary can be either for general purpose or trained for a specific application.

Our mathematical model for blind image restoration is stated as follows, where  images $X$ and $Y$ are converted to vectors $x$ and $y$, respectively:
\begin{eqnarray}
\left\{\begin{array}{l}
\min_{H,x} \| y - H x \|^2 + R(x) + \| \gamma \|_1 \\
H = \sum_{i} \gamma_i D_i.
\end{array}\right.  \label{oneDblurr}
\end{eqnarray}
where $R$ is a convex regularization term on image, $D = [D_i]$ is a tensor dictionary (vector of matrices) composed of basic blur kernel $D_i$,  and $\gamma =[\gamma_i]^T$ is a column vector of real numbers. The notation $H$ is overloaded here with slightly different meanings in Equations (\ref{blurreq}) and  (\ref{oneDblurr}).

The imposition of convex regularizations on $x$ and $\gamma$ facilitates an alternative approach to deriving the solution. Since each sub-problem, derived by fixing either $X$ or $H$,  is convex, the convergence of the approach is ensured. Nevertheless, the solution is still dependent on the initial guess of $H$ or $X$.  To note, modelling $H$ as a sparse representation of $D$ has an edge on guessing the initial $H$. If we have a dictionary of large number of atoms, a reasonable guess of $H$ to start an algorithm is a function that will contain only one atom in the dictionary (this corresponds to approximation of $H$ by its dominating atom). 

In this paper, we demonstrate our approach by constructing a dictionary of blurring patterns formed by the Kronecker product of two 1-D Gaussian functions of various scales. 
Our dictionary assumes that the unknown blur kernel is sparse with respect to a mixture of out-of-focus Gaussian type blurriness. Although our dictionary is not generic for all types of blurriness, the construction process can be used to derive dictionaries for other types of blurriness.  The initial guess of $H$ is derived  from estimation of the out-of-focus blurriness in images. 
The image is estimated by the variable splitting technique and the blur kernel is generated by the efficient proximal gradient method.  We also demonstrate our restoration results and compare the PSNR performance with other blind restoration approaches.

\noindent{\it{Notation.}}

Capital letters denote matrices and small letters denote vectors or scalars. 
The $vec$ operation transfers a matrix into a vector by stacking one vector underneath the other. The inverse of $vec$ is denoted as $vec^{-1}$.  If a capital letter is used to denote an image, the corresponding small letter will denote the vector obtained by applying $vec$  on the image. For example, if $X$ is an image, then $x$ is a vector with $x = vec(X)$. We use $\otimes$ to denote the Kronecker product. Applying $vec$ on both sides of $C = AXB$ yields 
\begin{equation}
vec(C) = vec(AXB) = (B^T \otimes A) vec(X).
\end{equation}

The rest of the paper is organized as follows. Section \ref{RelatedWork} reviews the related work.  Section \ref{PF} formulates the blind restoration problem, models a blurring kernel as a sparse representation of a dictionary, and devises the procedure to construct the dictionary.
In Section \ref{NA}, an alternating minimization algorithm is proposed for blind restoration. Some important steps of the algorithm are also discussed.
In Section \ref{ER}, we  will report the tests  conducted on monochrome and color image with various synthetic and real-life degradations and comparison of the results with other methods.
Section \ref{sec:con} contains our concluding remarks.

 

\section{Related Work} \label{RelatedWork}

Numerous approaches have been proposed to remove blurriness on observed images under various circumstances (the deblurring problem). They can be roughly categorized, according to whether the blur kernel  $H$ is spatially variant as well as how much information of the kernel is provided. Some of the relevant studies have been listed in Table \ref{table:RelatedWork} and categorized according to the problem addressed. Although a large volume of work has been  reported on deblurring problems,  the current trend appears to shift from non-blind to blind category.

The main technical challenge to resolve non-blind cases is imposition of regularizations on images.
On the other hand, the technical challenges for blind cases are the non-convexity of the problem, the determination of the blur kernel, and design of the regularizations on the blur kernel and the image.

Among all different forms of regularizations on images, the total variation (TV) regularization function and its variations have been widely used
\cite{Pan13, Chan98, QS08, Money08, Zhuo10, SB09, Cho09, YW08, Bect04, Nguyen11, Zunino09, Santosa96, Krishnan09}.  Statistical models on the gradients of natural images have been adopted for image prior \cite{Ferg06,QS08}. Sparse assumptions have been used to model representation coefficients for natural images in a transform domain \cite{Figueiredo05,Bect04} or in an image domain \cite{HZ11}. A counter-intuitive finding is reported in \cite{Ferg06} indicating that most cost functions for image prior prefer blurry images to sharp images, as blurry images have lower costs than sharp images. Attempt is made in \cite{Kris11} to achieve the lowest cost for true sharp image by introducing an $l_{1}/l_{2}$ function on images. 

The spatially variant cases are solved by dividing an image into blocks (of size depending on the supports of blur kernels) and then processing each block independently by a spatially invariant method.  Computational complexity is one of the main concerns for the spatially variant methods, because fast Fourier transformation (FFT) can be applied only to spatially invariant cases but not to spatially variant ones. Some fast algorithms based on variable splitting techniques and proximal point methods have also been proposed \cite{YW08,Bect04}. For example, the variable splitting technique is used in \cite{YW08} for the TV approach and in \cite{Krishnan09} for statistical modelling of image gradients. Another topic of concern is the removal of boundary effects incurred from dividing an image into blocks \cite{HJ12} such as the matting approach adopted in \cite{Nguyen11}.

As mentioned in \cite{Almeida10},  more the information of a blur kernel available for blind image restoration, better is the restoration performance achieved. Contrary to the regularizations on images, regularizations on blur kernels are more complicated because of the complex nature of the practical blurring processes. Depending on the sources of the blurriness of images, attempts have been made for regularizations on blur kernels. If the source of blurriness is motion of camera or objects, the blur kernel displays a trajectory with sharp edges, and can be modelled as a function of sparse support \cite{Ferg06,QS08,Xu12} or of a small total variation\cite{Money08}. On the other hand, if the source is due to out-of-focus in camera parameter, then a Gaussian type of smoothing is usually used to model the blurriness \cite{You99}.  A learning-based approach is also proposed to derive a blur kernel from an ensemble of training data \cite{Mis00}. Recently, Ji and Wang \cite{HJ12} analyzed the robustness issue of blind image restoration, and they reported that sometimes few errors on the estimated blind kernels can cause significant errors in the image reconstruction process. They introduced a model that can explicitly take into account the blur kernel estimation error  in the regularization process in order to obtain restoration results that are robust to modelling errors of blur kernels.

\begin{table}[h]
\begin{center}
\caption{Different deblurring problems and related works.} \label{table:RelatedWork}
\begin{tabular}{|c|c|c|}
\hline
 & Spatially invariant & Spatially variant\\
\hline 
& \cite{Pan13, Chan98, Ferg06, QS08, LW09}, & \cite{Jia07, CZF10, Zhang12, LFDF07},\\
blind & \cite{HZ11, Money08, SB09, Cho09}, &  \cite{Nguyen11, Ozkan94}\\
& \cite{Kris11, Cai09, Zunino09} &\\
\hline
 & Wiener filter, RL method & \\
non-blind & and \cite{LY07, Zhuo10, Rav05, Almeida10}, & \cite{NG1999, Nagy98}\\
& \cite{YW08, Bect04, Santosa96, Figueiredo05, Krishnan09}. & \\
\hline
\end{tabular}
\end{center}
\end{table}

\section{Problem Formulation} \label{PF}

Hereafter, we assume the blur kernel as a separable (block) circular convolution kernel. This assumption simplifies the derivations, brings computational efficiency (by employing two 1-D convolutions in place of a 2-D convolution), and maintains a generality in the proposed method. 
  
We use the following model for blind image restoration: 
\begin{eqnarray}
\left\{\begin{array}{l}
\min_{X, H_1, H_2} \| Y - H_1XH_2^T \|_F^2 + R(X) + \mu \sum_{i,j} |\alpha_i \beta_j|,  \label{formulation}\\
H_1  =  \sum_i \alpha_i D_i \label{H_1}   \\
H_2   =   \sum_j \beta_j \tilde D_j,   \label{H_2}
\end{array}\right.
\end{eqnarray}
where $H_1$ and $H_2^T$ are  (block) circulant matrices along the columns and rows of $X$, respectively;  $D_i$ and $\tilde D_j$ are circulant matrices representing basic blurring patterns for $H_1$ and $H_2$, respectively. The sum of each row of $H_i$ is normalized to $1$, so that $X$ and $Y$ have the same mean. Let ${\bf{1}}$ be an $n \times n$ image of all entries equal to $1$ and let $h_{i,k}$ be the $k$-column in $H_i$. Then, the mean of $H_1 {\bf{1}} H_2^T$ is $1$, as 
\begin{equation}
\frac{1}{n^2} \sum_{k,l=1}^n h_{1,k}^T {\bf{1}}_{k,l}  h_{2,l} = 1.
\end{equation}

We can make the dictionary for blur kernel explicit by taking $vec$ operation on matrices in the objective function in Equation (\ref{formulation}) to obtain
\begin{eqnarray}
\min_{x,H} \|y - H x \|^2 + R(x) + \sum_{i,j} |\alpha_i \beta_j|,   \label{oneD}
\end{eqnarray}
where 
\begin{eqnarray}
H  &= &  H_2 \otimes H_1 \\
&= & \sum_j \beta_j \tilde D_j   \otimes \sum_i \alpha_i D_i \\
& = & \sum_{i, j} \alpha_i \beta_j (\tilde D_j \otimes D_i). \label{Hexpression}
\end{eqnarray}
The term  $\sum_{i,j}|\alpha_i \beta_j|$ in Equation (\ref{formulation}) implies that the blur kernel $H$ is sparse with respect to the tensor dictionary, formed by a stack of matrix $\tilde D_j \otimes D_i$. $H$ is in the  vector space spanned by atoms $\tilde D_j \otimes D_i$, which are formed by applying the Kronecker product of the two circulant matrices, $\tilde D_j$ and $D_i $.
Note that a circulant matrix can be constructed from a 1-D sequence and the circulant matrix maintains the structure for convolution that can be efficiently implemented in frequency domain.

\subsection{Vector Space of Blur Kernel}

Since a blur kernel is usually a low-pass filter due to incorrect setting of camera parameters, we construct the vector space for the blur kernel $H$ by using low-pass circular patterns. We use Gaussian filters as our basis elements to compose the vector space for $H$ because Gaussian filters are not only separable, but also the most prevailing blurring operator for out-of-focus blurring distortions.

One-dimensional Gaussian functions of various standard deviations from $\sigma_{1}$ to $\sigma_{N}$ are uniformly quantized to obtain $N$ discrete sequences. Each sequence is then normalized so that the sum of the sequence is equal to $1$. The normalized sequence derived from the standard deviation $\sigma_i$ is then used to specify and generate the circulant matrix $G_i$. 
Let \begin{equation}
H_1 = \sum_{i=1}^N \alpha_i G_i, \label{H1r}
\end{equation} 
and 
\begin{equation}
H_2 = \sum_{i=1}^N \beta_i G_j. \label{H2r}
\end{equation} 
By using vector space representation, the circulant matrices $H_1$ and $H_2$ can be characterized by their respective coefficient vectors $\alpha = [\alpha_i]^T$ and $\beta = [\beta_i]^T$.
Since the sums of each row of $G_i$ and $H_i$ have been normalized, we have 
\begin{equation}
\sum_{i=1}^N \alpha_i= \sum_{i=1}^N \beta_i = 1.
\end{equation}

Substituting Equations (\ref{H1r}) and (\ref{H2r}) into the first term of the objective function in  problem (\ref{formulation}), we obtain
\begin{eqnarray}
Y&=&H_{1}XH_{2}^{T}+N\nonumber\\
&=&\sum_{i,j=1}^{N}\alpha_{i}\beta_{j}G_iXG_j^{T}+N,
\label{eq:13}
\end{eqnarray}
where $G_iXG_j^{T}$ is the blurred image of $X$, which is horizontally blurred by 1-D Gaussian of standard deviation $\sigma_j$ and vertically blurred by that of standard deviation $\sigma_i$; $N$ is the noise. 

The blur kernel $H$ is a linear combination of basic pattern $G_j \otimes G_i$ as
\begin{eqnarray}
H  &= &  H_2 \otimes H_1 \\
&= & \sum_{j=1}^N \beta_j G_j   \otimes \sum_{i=1}^N \alpha_i G_i \\
& = & \sum_{i, j=1}^N\alpha_i \beta_j (G_j \otimes G_i). \label{Hspace}
\end{eqnarray}
As each $G_j\otimes G_i$ matrix is of dimension $n^2 \times n^2$, $H$ is an $n^2 \times n^2$ matrix in the vector (sub)space of dimension $N^2$, spanned by the $N^2$ matrices $G_j \otimes G_i$.

\section{Numerical Algorithm} \label{NA}

To estimate the blur kernel $H$, estimation of its respective coefficients $\alpha$ and $\beta$ in the vector space is done, where $\alpha $ and $\beta$ denote the vectors of $\alpha_i$ and $\beta_j$ respectively. The proposed image restoration problem can now be re-expressed as
\begin{eqnarray}
\left\{\begin{array}{lll}
\min_{X,\alpha, \beta} \|Y - \sum_{i,j=1}^N \alpha_i \beta_j G_i X G_j^T \|_F^2 + R(X)\\
\,\,\,\,\,\,\,\,\,\,\,\,\,\,\,\,\,\,\,\,\,\,\,\,
 + \mu \sum_{i,j} |\alpha_i \beta_j|,  \\
\sum_{i} \alpha_i  =  1\\
\sum_{i} \beta_i  =  1.
\end{array}\right.
\end{eqnarray}
The problem can be solved by using a relaxation approach that alternatively estimate $X$ and ($\alpha, \beta$) considering the other variable fixed. If ($\alpha, \beta$) is fixed, $X$ can be obtained by solving the following sub-problem:
\begin{equation}
\min\limits_{X} \|Y-\sum_{i,j=1}^{N}\alpha_{i}\beta_{j}G_{i}XG_{j}^{T}\|_{F}^{2}+R(X).
\label{eq:14}
\end{equation}
Meanwhile, if $X$ is fixed, ($\alpha, \beta$) can be estimated by solving the following sub-problem:
\begin{equation}
\left\{\begin{array}{l}
\min\limits_{\alpha, \beta} \|Y-\sum_{i,j=1}^{N}\alpha_{i}\beta_{j}G_{i}XG_{j}^{T}\|_{F}^{2}+\mu\sum_{i,j} |\alpha_{i}\beta_{j}| \\
\sum_{i} \alpha_i  =  1\\
\sum_{i} \beta_i  =  1.
\label{eq:6}
\end{array}\right.
\end{equation}
If the regularization $R(X)$ is a convex function of $X$, then sub-problem (\ref{eq:14}) will be convex, too. Since sub-problems (\ref{eq:14}) and (\ref{eq:6}) are both convex, the convergence of alternative approach to  minimizers of $X$ and ($\alpha, \beta$) can be ensured. The algorithms to derive the minimizers of the above sub-problems are provided in the following subsections.

\subsection{Estimation of Image}

The algorithm for the sub-problem (\ref{eq:14}) can estimate image $X$.
By choosing the regularization on $X$ to be its total variation, sub-problem (\ref{eq:14}) becomes 
\begin{equation}
\min\limits_{X} \|Y-\sum_{i,j=1}^{N}\alpha_{i}\alpha_{j}G_{i}XG_{j}^{T}\|_{F}^{2}+\delta (\|D_1x\|_{1} + \|D_2x\|_1),
\label{eq:15}
\end{equation}
where $x$ is a vector form of $X$, $D_{1}x$ and $D_{2}x$ denote the vectors of the first-order discrete horizontal difference and vertical difference at each pixel of $X$, respectively; $\delta$ is a Lagrangian parameter.

Based on the work of Wang et al.\cite{YW08}, we use a variable splitting technique to estimate the image.  We replace
$\| D_1 x\|_1$ by $\| v_1 \|_1 + \frac{\gamma}{2} \| D_1 x - v_1 \|_2^2$ and 
$\| D_2 x\|_1$ with $\| v_2 \|_1 + \frac{\gamma}{2} \| D_2 x - v_2 \|_2^2$, by introducing new variable vectors $v_1$ and $v_2$.
As $\gamma \rightarrow \infty$, it is clear that 
\begin{equation}
\| v_i \|_1 + \frac{\gamma}{2} \| D_i x - v_i \|_2^2 \rightarrow \| D_i  x\|_1, 
\end{equation}
for $i = 1, 2$.
Sub-problem (\ref{eq:14}) can now be re-written as 
\begin{equation}
\min\limits_{X, v} \|Y-\sum_{i,j=1}^{N}\alpha_{i}\beta_{j}G_{i}XG_{j}^{T}\|_{F}^{2}+
\delta\sum_{i = 1}^2 (\|v_{i}\|_{1}+\frac{\gamma}{2}\|v_{i}-D_{i}x\|_{2}^{2}),
\label{eq:21}
\end{equation} 
where $v =[ v_1^T  v_2^T]^T$.
The solution of the above equation converges to that of sub-problem (\ref{eq:14}) as $\gamma\rightarrow\infty$. The variable splitting technique is extremely efficient, because when either of the two variables in Equation (\ref{eq:21}) is fixed, minimizing the equation with respect to the other has a closed-form solution. The overall convergence of this minimization algorithm is well analyzed in \cite{YW08}.

For a fixed $X$, variables $v_1$ and $v_2$ can be derived separately by solving 
\begin{equation}
\min\limits_{v_{i}} \|v_{i}\|_{1}+\frac{\gamma}{2}\|v_{i}-D_{i}x\|_{2}^{2}.
\label{eq:22}
\end{equation}
The $l_1$-minimizer is given by the following shrinkage formula:
\begin{equation}
v_{i}=max\bigg\{|D_{i}x|-\frac{1}{\gamma}, 0\bigg\}sign(D_{i}x),
\label{eq:23}
\end{equation}
where all operations are done component-wise. On the other hand, for a fixed $v$, the minimizer $X$ can be derived by  
\begin{equation}
\min\limits_{x} \|y - \sum_{i,j=1}^N \alpha_i \beta_j (G_j \otimes G_i) x\|_{F}^{2}+\frac{\delta\gamma}{2}\sum_{i=1}^2\|v_{i}-D_{i}x\|_{2}^{2},
\label{eq:24}
\end{equation} 
where the first term is obtained by taking $vec$ operation on the first term of sub-problem (\ref{eq:14}). Let $A = \sum_{i,j=1}^N \alpha_i \beta_j (G_j \otimes G_i)$. 
Taking the partial derivative of Equation (\ref{eq:24}) with respect to $x$ and setting the resultant to zero, we obtain the following equation for the minimizer $x$:
\begin{equation}
\bigg(\sum_{i=1}^2D_{i}^{T}D_{i}+\frac{2}{\delta\gamma}A^{T}A\bigg)x=\sum_{i=1}^2D_{i}^{T}v_{i}+\frac{2}{\delta\gamma}A^{T}y.
\label{eq:25}
\end{equation} 
If $X$ is an $n \times n$ matrix, then the size of $A$ will be $n^{2}\times n^{2}$, which becomes cumbersome for  direct computation of Equation (\ref{eq:25}) as we need to inverse a huge matrix. Thus, it is suggested in \cite{YW08} to solve it by using the FFT.

If we explore the (block) circulant matrix structure of $D_1$, $D_2$, and each term in $A$ \cite{NG1999}, all matrix multiplications in Equation (\ref{eq:25}) are convolution operations. Let 
$\mathcal{F}_1(C)$ denote the 1-D Fourier transform of the generating sequence of circulant matrix $C$.
Through convolution theorem of Fourier transform, Equation (\ref{eq:25}) can be written as
\begin{equation}
x=\mathcal{F}_1^{-1}\bigg(\frac{\sum_{i=1}^{2}(\overline{\mathcal{F}_1(D_{i})} \circ \mathcal{F}_1(v_i) ) + \frac{2}{\delta\gamma} \overline{\mathcal{F}_1(A)} \circ \mathcal{F}_1(y)} {\sum_{i=1}^{2}(\overline{\mathcal{F}_1(D_{i})} \circ \mathcal{F}_1(D_{i}) ) + \frac{2}{\delta\gamma} \overline{\mathcal{F}_1(A)} \circ \mathcal{F}_1(A)}\bigg),
\label{eq:26}
\end{equation}
where $\mathcal{F}_1$ and $\mathcal{F}_1^{-1}$ denote the forward and inverse 1-D Fourier transform, respectively, $\circ$ denotes component-wise multiplication, and $\overline{\mathcal{F}_1(C)}$ is the complex conjugate of $\mathcal{F}_1(C)$. Since only the variables $v_1$ and $v_2$ of Equation (\ref{eq:26}) are changed during each iteration, computation loads can be reduced by computing the FFT of all the other variables in advance.

\subsection{Estimation of the Blur Kernel}

If we let $Z_{i,j} = G_{i}XG_{j}^{T}$, the sub-problem (\ref{eq:6}) can be expressed as
\begin{equation}
\left\{\begin{array}{l}
\min\limits_{\alpha, \beta} \|Y-\sum_{i,j=1}^{N}\alpha_{i}\beta_{j}Z_{i,j}\|_{F}^{2}+\mu \sum_{i,j} |\alpha_{i}\beta_{j}|  \\
\sum_{i} \alpha_i  =  1\\
\sum_{i} \beta_i  =  1.
\end{array}\right. \label{estbr}
\end{equation}
Let $U$ be the rank one matrix $\alpha \beta^T$, where $\alpha$ and $\beta$ are vectors of $\alpha_i$ and $\beta_i$, respectively; and let $u = vec (U)$. We have $\sum_{i,j} |\alpha_{i}\beta_{j}| = \|u \|_1$. In addition, 
If  $vec$ is applied on the first term of problem (\ref{estbr}), we obtain
\begin{equation}
 \|y-Zu\|_{2}^{2},
\label{eq:28}
\end{equation} 
where $y = vec(Y)$ and $Z$ is a matrix with columns corresponding to the vectors derived by $vec(Z_{i,j})$. Problem (\ref{estbr}) can now be re-expressed as
\begin{equation}
\left\{\begin{array}{l}
\min\limits_{\alpha, \beta} \frac{1}{2}\|y-Zu\|_{2}^{2}+ \mu \|u\|_1  \\
u = vec(U) \\
U = \alpha \beta^T  \text{ //$U$ is a rank one matrix.} \\
\sum_{i} \alpha_i  =  1\\
\sum_{i} \beta_i  =  1.
\end{array}\right.  \label{estbr1}
\end{equation}
First, all constraints of the above problem are ignored, and the proximal gradient method is applied to derive $u$ by solving the objective function.
Then, the constraints are imposed on $u$ so that the resultant matrix $U = vec^{-1}(u)$ becomes a rank one matrix with $\sum_i \alpha_i = \sum_i \beta_i = 1$.

The first term in the objective of problem (\ref{estbr1}) is differentiable and the second term is indifferentiable.  
The solver of the proximal gradient method is thus
\begin{equation}
u^{k+1}:=\textbf{prox}_{\mu^k \|.\|_1}(u^{k}-\mu^k\bigtriangledown f(u^{k})),
\label{eq:29}
\end{equation} 
where 
\begin{equation}
\bigtriangledown f(u^{k})=Z^{T}Zu^{k}-Z^{T}y,
\label{eq:30}
\end{equation} 
and $\textbf{prox}_{\mu^k \|.\|_1}$ is soft thresholding and $\mu^k$ is a step size. The step size $\mu^{k}$ is determined by backtracking line search process.  The proximal gradient method to derive initial $u$ is detailed in the following.

{
\hspace{-0.5cm}
\begin{tabular}[h]{@{}p{8.8cm}@{}}
\hline
{\bf given:}
$u^{k}$, $\mu^{0}$, and parameter $\beta\in$(0, 1).
\\Let $\mu:=\mu^{0}$.
\\{\bf repeat}
\\$\qquad$1. Let $z:=\textbf{prox}_{\mu\|.\|_1}(u^{k}-\mu\bigtriangledown f(u^{k}))$.
\\$\qquad$2. $\textbf{break if}$ $f(z)\leq\Hat{f}_{\mu}(z,u^{k})$.  //defined in Equation (\ref{eq:31})
\\$\qquad$3. Update $\mu:=\beta\mu$. Go to 1.
\\{\bf return}
$\mu:=\mu^0$, $u^{k}:= z$ 
\\
\hline
\\
\end{tabular}
}
The typical value for the line search parameter $\beta$ is set at $0.5$, and the $\Hat{f}_{\mu}$ in line 2 is the stopping criterion given as 
\begin{equation}
\Hat{f}_{\mu}(x,y)=f(y)+\bigtriangledown f(y)^{T}(x-y)+\frac{1}{2\mu}\|x-y\|_{2}^{2},
\label{eq:31}
\end{equation} 
with $\mu>$0. This proximal gradient method needs to be evaluated many times in the minimization process until the optimal $u$ is reached (in line {\bf{return}}).

Let $\hat u$ be the result of the proximal gradient method. Then, we apply $vec^{-1}$ on $\hat u$ to obtain the matrix $\hat{U}$. Since the objective matrix $U$ is a rank one matrix, singular value decomposition is applied on $\hat{U} $ to obtain
\begin{equation}
\hat U = S\Sigma V^T,
\end{equation}
where the singular values are arranged in a non-increasing order\footnote{If $H_1 = H_2$ (the horizontal blurring and the vertical blurring are performed by the same kernel), we can project $\hat{U}$ into the vector space of symmetric matrix to obtain its projection $\frac{\hat U +\hat U^T}{2}$, which is then a rank one symmetric matrix. }.
The optimum rank one approximation of $\hat U$ is the matrix of $\sigma s v^T$, where $s$ is the first column in $S$, $v^T$ is the first row in $V^T$ and $\sigma$ is the largest eigenvalue in $\Sigma$.  Since 
\begin{equation}
U = \sigma s v^T = \alpha \beta^T, \label{Umatrix}
\end{equation}
we can factor $\sigma = a b$ such that
\begin{equation}
\sum_i \alpha_i = a \sum_i s_i = 1
\end{equation}
and 
\begin{equation}
\sum_i \beta_i = b \sum_i v_i = 1.
\end{equation}
Multiplying both sides of the last equation by $a$, we have
\begin{equation}
a = \sigma \sum_i v_i . \label{determinea}
\end{equation}
Note that if we know $H_1 = H_2$, then $\alpha = \beta$ and $a = b = \sqrt{\sigma}$.

\subsection{Algorithm}


\begin{figure}[!ht]
\begin{center}
  \includegraphics[width=0.9\linewidth]{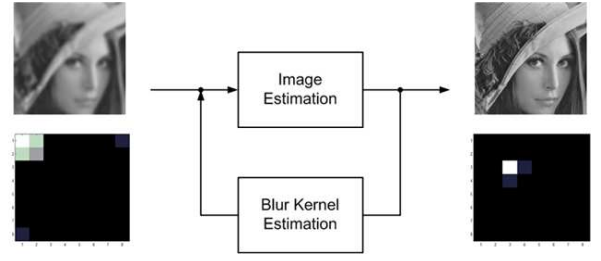}
  \caption{Overview of the proposed algorithm. The top left subimage is the observed blurred image. The bottom left subimage is formed from the coefficients of the initial blur kernel, with the horizontal and vertical axes corresponding to the initial $\alpha$ and $\beta$ vector, respectively.
Image and blur kernel are iteratively refined until the convergence is reached, as shown in the right subimages.}\label{fig:algorithmflow}
\end{center}
\end{figure}

Figure $\ref{fig:algorithmflow}$ displays an overview of the proposed alternative optimization approach, with the stepwise algorithm given in the following table.

\hspace{-0.5cm}
\begin{tabular}[h]{@{}p{8.8cm}@{}}

\hline
{\bf Input:}
Blurry image $Y$; initial $\alpha$ and $\beta$;  parameters $\gamma <p$, where $p$ is a given constant; $\delta$ in Equation (\ref{eq:15}); a dictionary of $N^2$ basic patterns; and the maximum number of iterations $M$. 
\\Let $X=Y$.  // Image is initialized as $Y$.
\\1. If $\gamma < p$, then 
\\$\qquad${\bf repeat}
\\$\qquad$2. Estimate image $X = vec^{-1}(x)$ according to Equation 
\\$\qquad\quad$(\ref{eq:26}).
\\$\qquad$3. Estimate $\alpha$ and $\beta$ by solving problem (\ref{estbr1}):
\\$\qquad\qquad$3.1. Update $u$ by proximal gradient method, based 
\\$\qquad\qquad\quad$on Equation (\ref{eq:29}).
\\$\qquad\qquad$3.2. Derive the rank one approximation $U$ of 
\\$\qquad\qquad\quad$$vec^{-1}(u)$.
\\$\qquad\qquad$3.3. Determine $\alpha$ and $\beta$ from SVD of $U$, based on 
\\$\qquad\qquad\quad$Equation (\ref{determinea}).
\\$\qquad$4. $\textbf{break if}$ maximum iteration number $M$ is reached.
\\5. Increase $\gamma$. Go to 1.
\\6. Derive $H$ from the coefficients $\alpha$ and $\beta$, based on Equation 
\\$\quad$(\ref{Hexpression}).
\\{\bf return}
Image $X$ and blur kernel $H$.
\\
\hline
\\
\end{tabular}

The outer loop, composed of steps 1 and 5, of this algorithm increase the parameter $\gamma$ for the variable splitting method. Starting with a small $\gamma$ value, our algorithm gradually increase its value to reach a given constant $p$. The inner loop (including steps 2, 3, and 4) alternatively estimate image $X$ and the blur kernel coefficients $\alpha$ and $\beta$. In each step of estimation in the inner loop, the minimum of a convex function is found; therefore,  the inner loop always show convergence. Without degrading the overall performance of our algorithm, we set a maximum number of iterations to enforce the inner loop to stop before it reaches the convergence.

Because of non-convexity of the blind image restoration problem, initial guess of $H$ is surely important for the final restoration result. Empirically,  the current method can yield a good restoration image if the prior information about the blur kernel can be well estimated. Degrading the initial guess could result in a slow-paced convergence.
Since $H$ is assumed to be sparse, if the number of atoms in a dictionary is large enough, a good initial guess of $H$ can be the one sparse function that  approximates $H$ by one atom in the dictionary. For an image subjected to Gaussian-type out-of-focus blurriness, there are algorithms that can reliably estimate the dominating Gaussian blur kernel on the image. For other dictionaries, where it is difficult to identify the dominating atom, the initial guessing of $H$  cannot be easily identified, an exhaustive approach based on trying on each atom as the initial guess of $H$ can be applied.

\section{Experimental Results} \label{ER}

Here we intend to show that our method can be effectively performed on different images with degradations by various blur kernels in both noise-less and noisy environments.

\noindent{A. Dictionary Design and Initial Guess of H}:

In this part, a detailed description of our dictionary design has been provided.   We chose $N$ scaled 1-D Gaussian  function, $G(\sigma)$, as the building block for the 2D-atoms of our dictionary. The dictionary has $N^2$ atoms, each is of the form $G(\sigma_i) \otimes G(\sigma_j)$, with $i, j = 1, \cdots, N$. Increasing $N$ can increase the restoration performance, but also increase the computational complexity at the same time. To achieve an optimal balance between performance and computational efficiency, the value of $N$ is empirically determined to be eight. The standard deviations  of 1-D Gaussian functions corresponding to the eight number are set from $0.5$, $1$, $\cdots$, $3.5$, $4$. For memory consideration, we processed our method by blocks, where an image is divided into blocks of size $32 \times 32$.  Instead of processing each $32 \times 32$ block directly, to avoid the boundary artefact, we processed on a larger block (called processing block) of size $96 \times 96$ that embeds a $32 \times 32$ block in the center and then took the center $32 \times 32$ block in the result. Therefore, our dictionary has $64$ 2-D patterns, each of dimension $96^2 \times 96^2$.

We use the all-focused method in \cite{Zhang12} to estimate the out-of-focus blur on an image. The blurriness of step edges is modelled as  the convolution of a 2-D Gaussian function, approximating the point spread function of a camera. From the horizontal and vertical blurriness, the standard deviation of the Gaussian function is derived.  Let $\sigma_{i_0}$ and $\sigma_{j_0}$ be the estimated horizontal and vertical standard deviations, respectively.  Since our dictionary has only $64$ atoms, this number is not large enough to approximate a blur kernel by one sparse function of atom, we thus used the basis pursuit denoising  algorithm (BPDN) \cite{Berg07} to derive the approximation of $G(\sigma_{i_0}) \otimes G(\sigma_{j_0})$. From Equations (\ref{H1r}) and (\ref{H2r}),  the following optimization problem is solved by BPDN:
\begin{eqnarray}
\left\{\begin{array}{l}
\min \|\alpha\|_{1}  + \|\beta\|_1 \\
\|G(\sigma_{i_0}) -  \sum_{i=1}^N \alpha_i G(\sigma_i)\|_2 \leq \tau \\
\|G(\sigma_{j_0}) -  \sum_{i=1}^N \beta_i G(\sigma_i)\|_2 \leq \tau,  
\label{eq:19}
\end{array}\right.
\end{eqnarray}
where $\tau$ is given as the error bound. Let the index set of non-zero coefficients of  $\alpha$ be $I_1$ and that of non-zero coefficients of $\beta$ be $I_2$. Then, the initial guess of $H$ is $H_2^0\otimes H_1^0$, where
$H_1^0 = \sum_{i \in I_1} \alpha_i G(\sigma_i)$ and $H_2^0 = \sum_{i \in I_2} \beta_i G(\sigma_i)$.

\noindent{B. Comparisons and Algorithm Parameters}:

We further compare our results with two other deblurring methods, viz. the matlab built-in function $deconvblind$ and the method proposed by Krishnan $et$ $al.$ \cite{Kris11}. The $deconvblind$ deconvolves a blurred image by using the maximum likelihood algorithm.  The $deconvblind$ has several optional parameters, e.g. number of iterations. $deconvblind$ is quite fast, but the deblurred results in terms of PSNR as well as visual quality are often unsatisfactory, even with several iterations. On the other hands, Krishnan's method has many manually selected parameters. The two parameters that we chose are different  from their default  values. The chosen parameters are $\lambda$ (the regularizing parameter), which ranged from $60$ to $100$, and the iteration number, which is set to $20$.

\begin{figure}[t]
\begin{center}
\begin{tabular}{cc}
\includegraphics[width=0.4\linewidth]{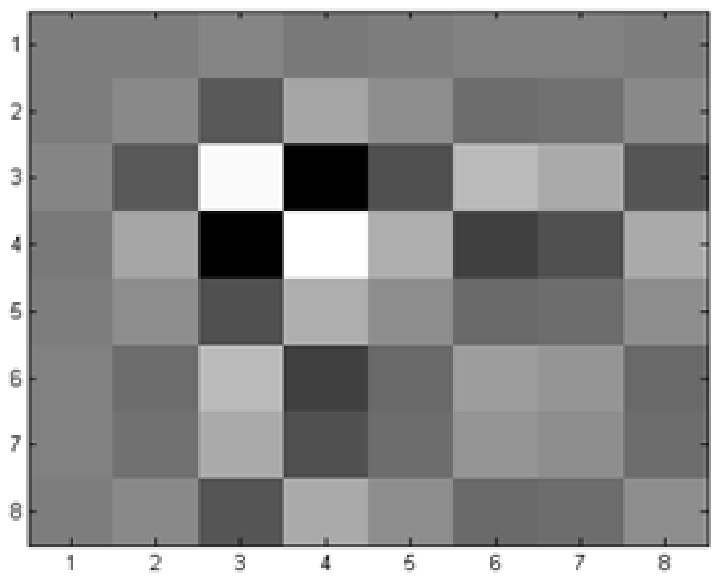}&
\includegraphics[width=0.4\linewidth]{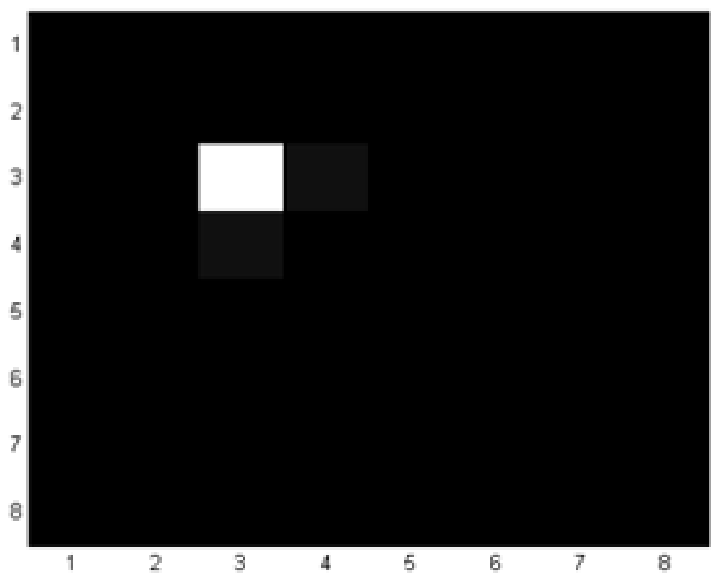}\\
kernel 1 & kernel 2\\
\includegraphics[width=0.4\linewidth]{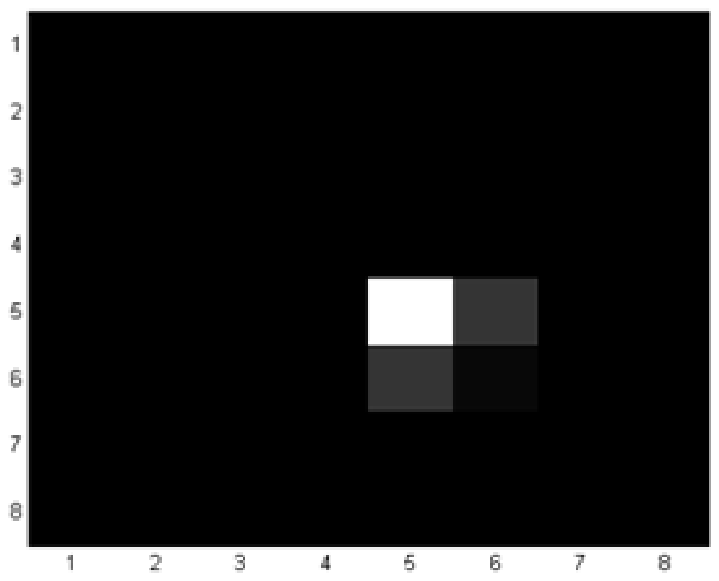} &
\includegraphics[width=0.4\linewidth]{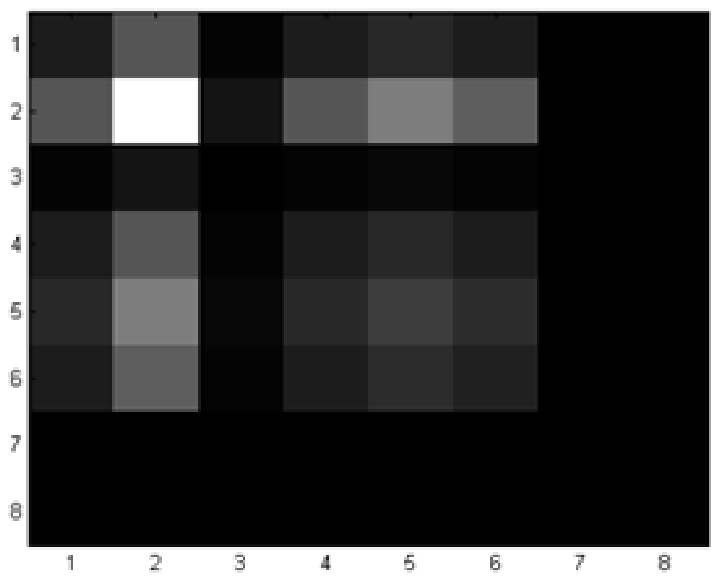}\\
kernel 3 & kernel 4\\
\end{tabular}
\caption{Representing the kernels in Table $\ref{table:PSF}$ with respect to our dictionary. The sub-figures are the absolute values of coefficients $\alpha\beta^{T}$. Kernels $2$ and $3$ are sparse.} \label{fig:psd_coeff}
\end{center}
\end{figure}

We used the four $512\times 512$ grayscale images, $Lena $(Img01), $Cameraman$(Img02), $House$(Img03) and $Mandrill$(Img04), and four blur kernels in \cite{Dan11} as our benchmark 
for the first two experiments. In total, we have $16$ blurred images.
The point spread function (PSF) of each blur kernel $H$ is normalized so that $\sum_{i}h_i=1$. The numerical values of the kernels are given in Table $\ref{table:PSF}$. Figure $\ref{fig:psd_coeff}$ shows the maps of the absolute values of coefficients $\alpha\beta^{T}$, defined in Equations (\ref{estbr1}) and (\ref{Umatrix}), for each of the four blur kernels. The darker a pixel is, smaller the value of the pixel has. As shown in the figure, kernels 2 and 3 are sparse with respect to our dictionary, and Kernel 4 is almost sparse.

\begin{table}[t]
\begin{center}
\caption{Th PSFs of blur kernels used for our comparisons.}
\label{table:PSF}
\begin{tabular}{|c|c|}

\hline
kernel & PSF\\
\hline 
1 & disk with radius=5\\
\hline
2 & $H=h_{0}h_{0}^T$, $h_{0}^T$=[1 9 36 84 126 126 84 36 9 1]\\
\hline
3 & Gaussian with $\sigma$=2.6\\
\hline
4 & $H$=1/(1+$x_{1}^{2}$+$x_{2}^{2}$), $x_{1}$, $x_{2}$=-7,$\ldots$,7\\
\hline
\end{tabular}
\end{center}
\end{table}

To make a quantitative comparison, we use the peak signal-to-noise ratio (PSNR) and sum-of-squared differences (SSD) to measure the accuracy of the deblurred image and the estimated PSF, respectively. The experimental results for all the compared methods are derived  based on the same initial kernel, as described in Part A of this section, and the same  number of iterations ($20$ runs for all cases). The step size used in our proximal gradient method is chosen with $\mu \in [5e^{-11}, 5e^{-13}]$ and the value of $\gamma$ is from $1$ to $100$ (the value of $p$ in our algorithm) and $\delta$ is between $[10^{-3},10^{-1}]$.  

\begin{table*}[t]
\begin{center}
\caption{Comparison of the output PSNR [dB] and SSD of the three deconvolution methods on $16$ images.}
\begin{tabular}{|c|c|c|c|c|c|c|c|c|c|}
\hline
\multicolumn{2}{|c|}{} & \multicolumn{2}{|c|}{Kernel 1} & \multicolumn{2}{|c|}{Kernel 2} & \multicolumn{2}{|c|}{Kernel 3} & \multicolumn{2}{|c|}{Kernel 4}\\\cline{3-10}
\multicolumn{2}{|c|}{} & PSNR & SSD & PSNR & SSD & PSNR & SSD & PSNR & SSD\\
\hline 
& deconvblind & \textbf{26.54} & 0.0959 & \textbf{29.32} & 0.0319 & 27.07 & 0.0018 & 30.83 & 0.0283\\
Img01 & Krishnan et al. & 22.42 & \textbf{0.0065} & 23.41 & \textbf{0.0087} & 24.05 & 0.0170 & 29.30 & \textbf{0.0027}\\
& Our method & 26.48 & 0.0948 & 28.56 & 0.0104 & \textbf{28.58} & \textbf{0.0004} & \textbf{31.20} & 0.0030\\
\hline

& deconvblind & \textbf{27.27} & 0.0959 & 31.18 & 0.0319 & 28.06 & 0.0023 & 32.29 & 0.0280\\
Img02 & Krishnan et al. & 22.31 & \textbf{0.0112} & 28.61 & 0.0189 & \textbf{29.87} & 0.0237 & \textbf{33.01} & \textbf{0.0027}\\
& Our method & 26.95 & 0.0971 & \textbf{31.28} & \textbf{0.0104} & 29.35 & \textbf{0.0004} & 31.07 & 0.0030\\
\hline

& deconvblind & 31.94 & 0.0959 & 35.39 & 0.0319 & 32.57 & 0.0015 & 28.75 & 0.0289\\
Img03 & Krishnan et al. & 25.09 & \textbf{0.0059} & 27.53 & 0.0122 & 28.02 & 0.0071 & \textbf{35.06} & 0.0043\\
& Our method & \textbf{32.51} & 0.0963 & \textbf{35.54} & \textbf{0.0106} & \textbf{32.65} & \textbf{0.0004} & 34.70 & \textbf{0.0032}\\
\hline

& deconvblind & 21.92 & 0.0958 & \textbf{25.32} & 0.0318 & 22.64 & 0.0024 & 27.65 & 0.028\\
Img04 & Krishnan et al. & 19.79 & \textbf{0.0109} & 21.66 & \textbf{0.0044} & 19.18 & 0.0315 & 24.84 & 0.0071\\
& Our method & \textbf{21.97} & 0.0955 & 24.57 & 0.0104 & \textbf{23.56} & \textbf{0.0004} & \textbf{29.24} & \textbf{0.0046}\\
\hline
\end{tabular}
\label{table:comparison}
\end{center}
\end{table*}
\begin{table*}[t]
\begin{center}
\caption{Comparison of the output PSNR [dB] of the three deconvolution methods with images stained by $30$ dB additive white Gaussian noise.}
\begin{tabular}{|c|c|c|c|c|c|c|c|c|c|}
\hline
\multicolumn{2}{|c|}{Each with} & \multicolumn{2}{|c|}{Kernel 1} & \multicolumn{2}{|c|}{Kernel 2} & \multicolumn{2}{|c|}{Kernel 3} & \multicolumn{2}{|c|}{Kernel 4}\\\cline{3-10}
\multicolumn{2}{|c|}{noise 30dB} & PSNR & SSD & PSNR & SSD & PSNR & SSD & PSNR & SSD\\
\hline 
& deconvblind & \textbf{24.56} & 0.1123 & 25.97 & 0.0319 & 19.67 & 0.0018 & 24.33 & 0.0858\\
Img01 & Krishnan et al. & 22.32 & \textbf{0.0089} & 23.22 & \textbf{0.0088} & 23.86 & 0.0166 & \textbf{29.54} & 0.0055\\
& Our method & 23.78 & 0.1022 & \textbf{26.51} & 0.0140 & \textbf{27.01} & \textbf{0.0008} & 28.11 & \textbf{0.0048}\\
\hline

& deconvblind & \textbf{25.73} & 0.1121 & 28.20 & 0.0319 & 19.26 & 0.0023 & 25.20 & 0.0857\\
Img02 & Krishnan et al. & 22.25 & \textbf{0.0119} & 28.43 & 0.0188 & \textbf{29.68} & 0.0247 & \textbf{33.51} & 0.0072\\
& Our method & 25.63 & 0.1046 & \textbf{29.68} & \textbf{0.0148} & 29.00 & \textbf{0.0007} & 27.29 & \textbf{0.0043}\\
\hline

& deconvblind & 24.93 & 0.1127 & 24.85 & 0.0319 & 22.10 & 0.0015 & 25.13 & 0.0858\\
Img03 & Krishnan et al. & 24.98 & \textbf{0.0071} & 27.22 & \textbf{0.0022} & 27.72 & 0.0055 & \textbf{34.38} & 0.0045\\
& Our method & \textbf{30.98} & 0.1029 & \textbf{31.86} & 0.0141 & \textbf{31.28} & \textbf{0.0007} & 30.56 & \textbf{0.0037}\\
\hline

& deconvblind & \textbf{21.71} & 0.1119 & 23.41 & 0.0318 & 15.55 & 0.0024 & 24.34 & 0.0857\\
Img04 & Krishnan et al. & 19.56 & \textbf{0.0160} & 21.62 & \textbf{0.0062} & 19.03 & 0.0329 & 25.24 & \textbf{0.0046}\\
& Our method & 21.06 & 0.1028 & \textbf{23.79} & 0.0135 & \textbf{21.03} & \textbf{0.0007} & \textbf{25.76} & 0.0049\\
\hline
\end{tabular}
\label{table:NoiseComparison}
\end{center}
\end{table*}

\begin{figure*}[t]
\begin{center}
\mbox{
{\includegraphics[width=0.2\textwidth]{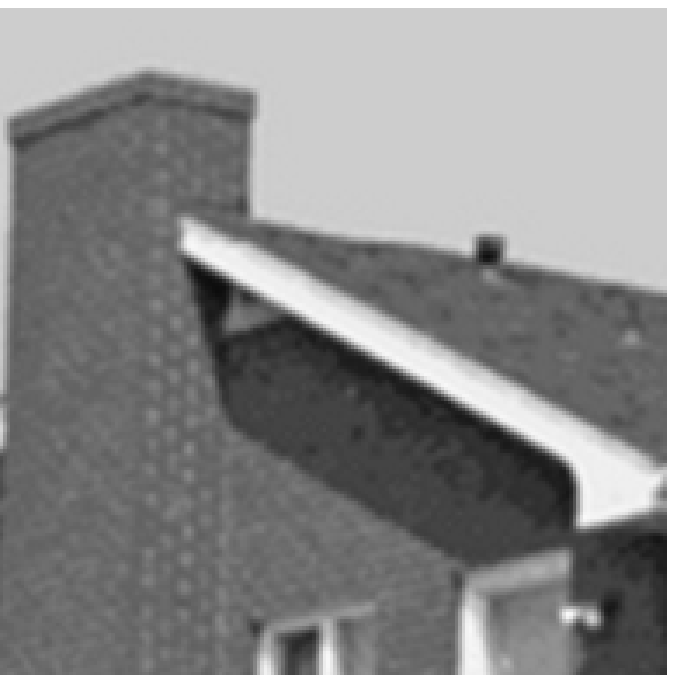}}
{\includegraphics[width=0.2\textwidth]{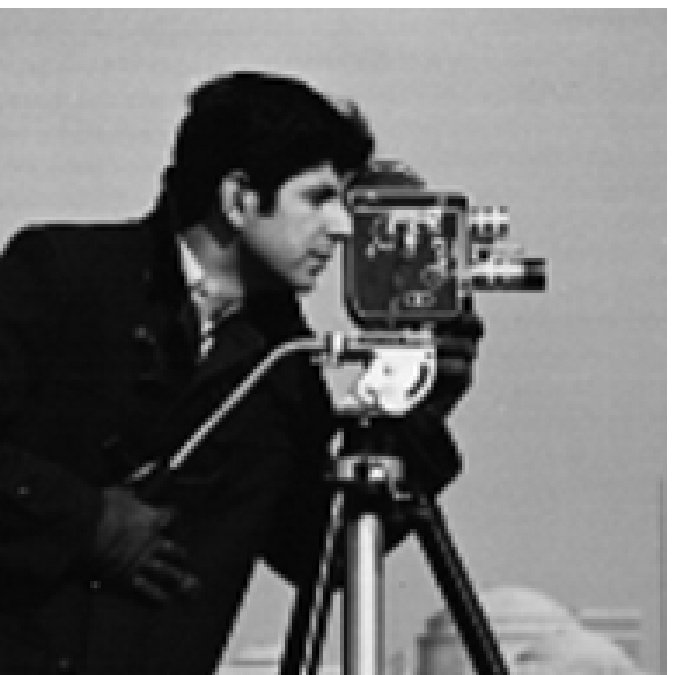}}
{\includegraphics[width=0.2\textwidth]{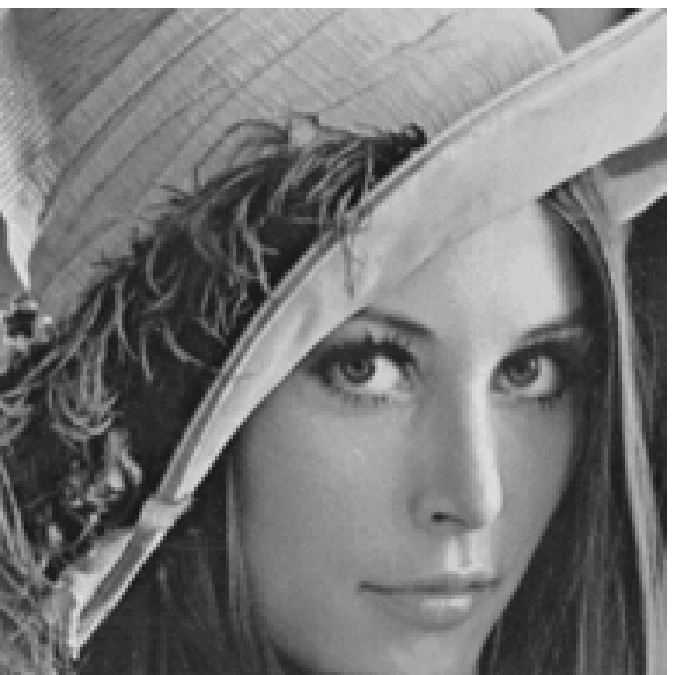}}
{\includegraphics[width=0.2\textwidth]{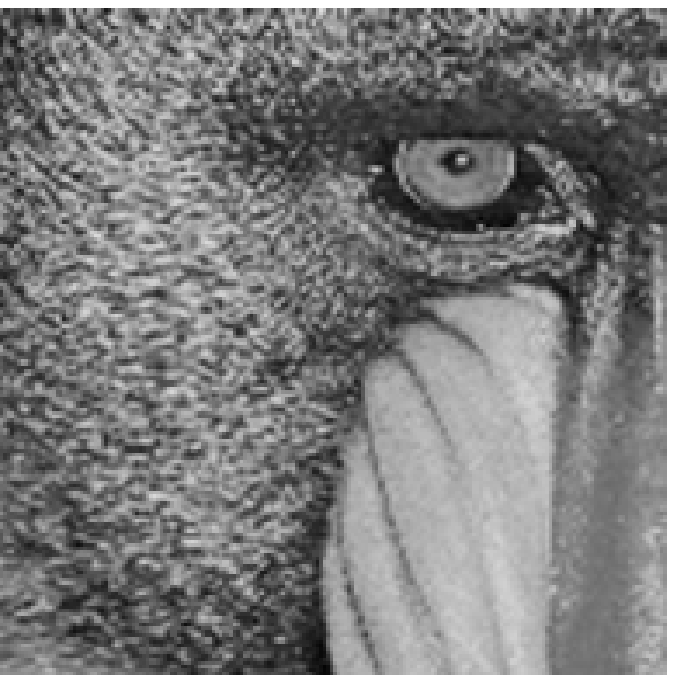}}
}
\mbox{
{\includegraphics[width=0.2\textwidth]{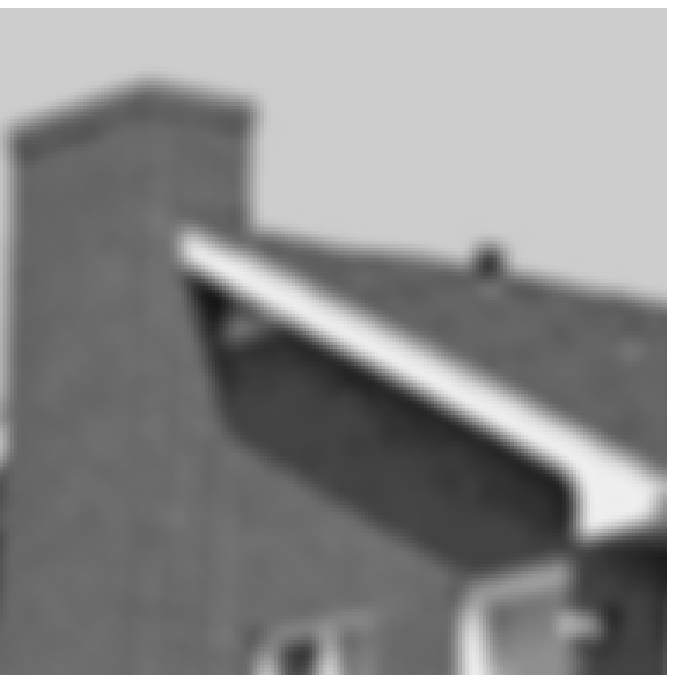}}
{\includegraphics[width=0.2\textwidth]{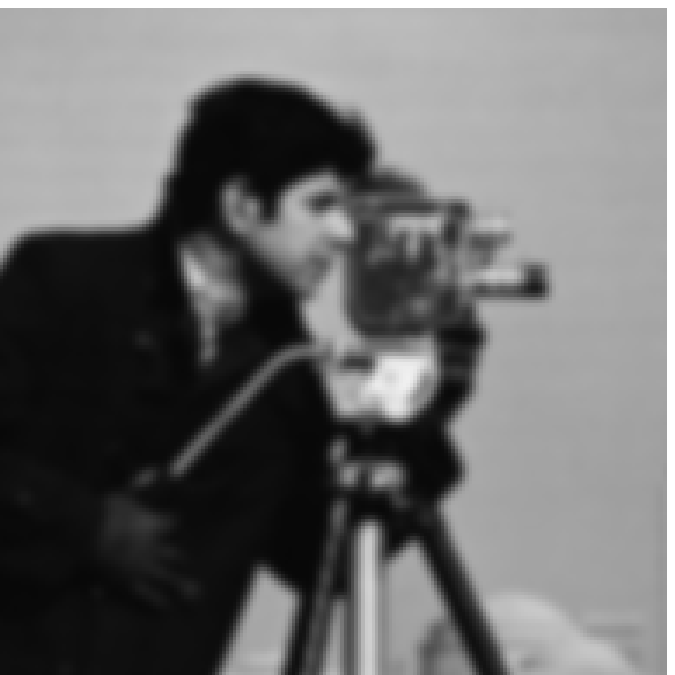}}
{\includegraphics[width=0.2\textwidth]{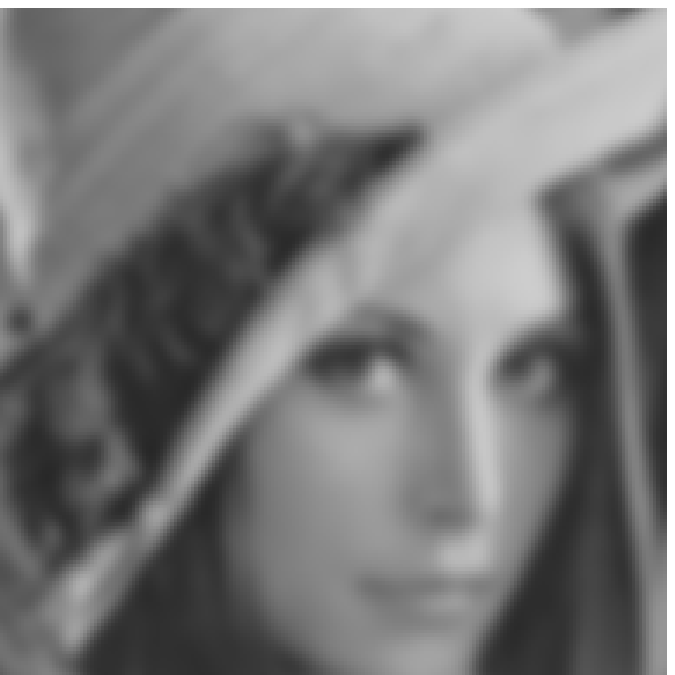}}
{\includegraphics[width=0.2\textwidth]{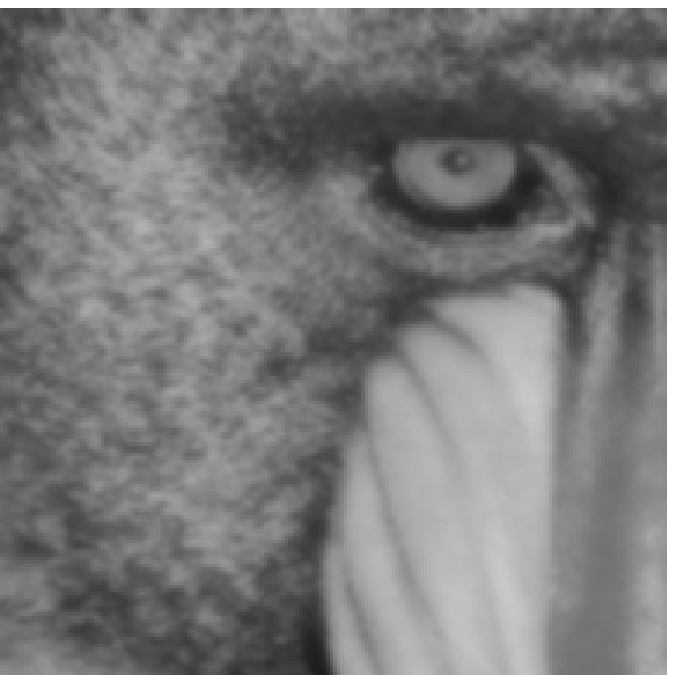}}
}
\mbox{
{\includegraphics[width=0.2\textwidth]{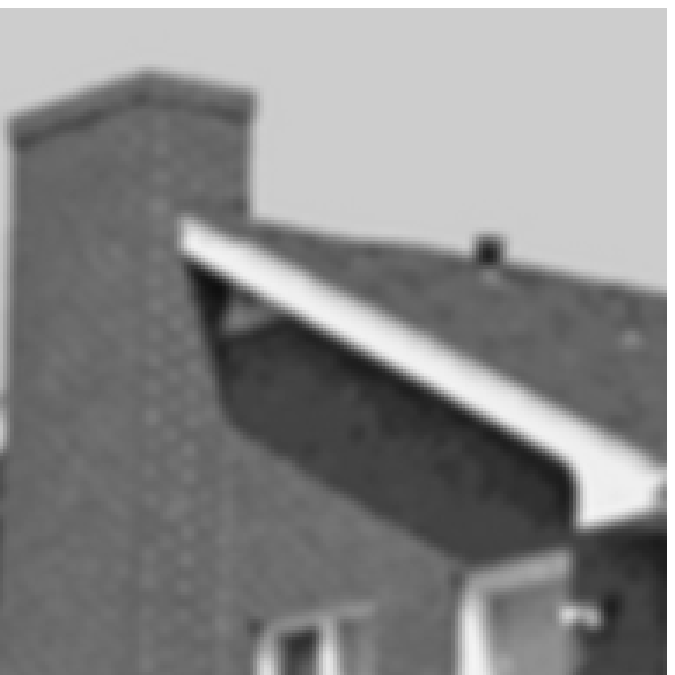}}
{\includegraphics[width=0.2\textwidth]{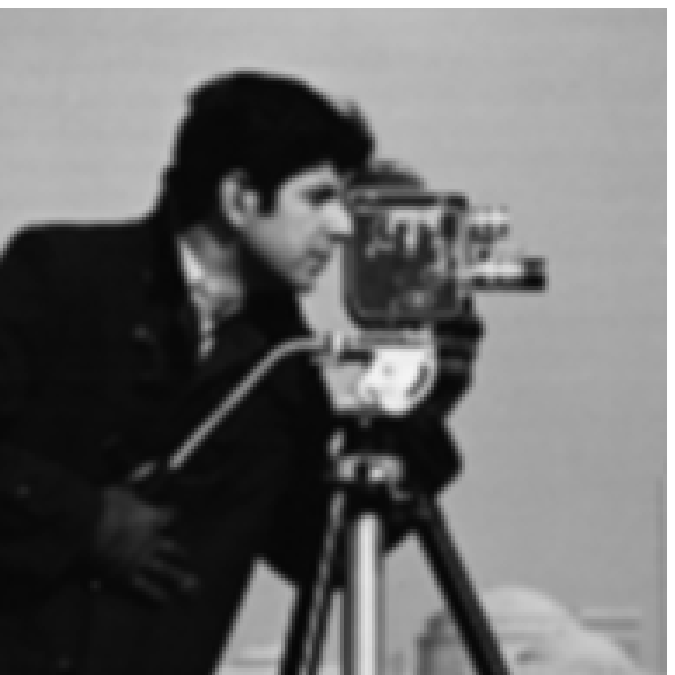}}
{\includegraphics[width=0.2\textwidth]{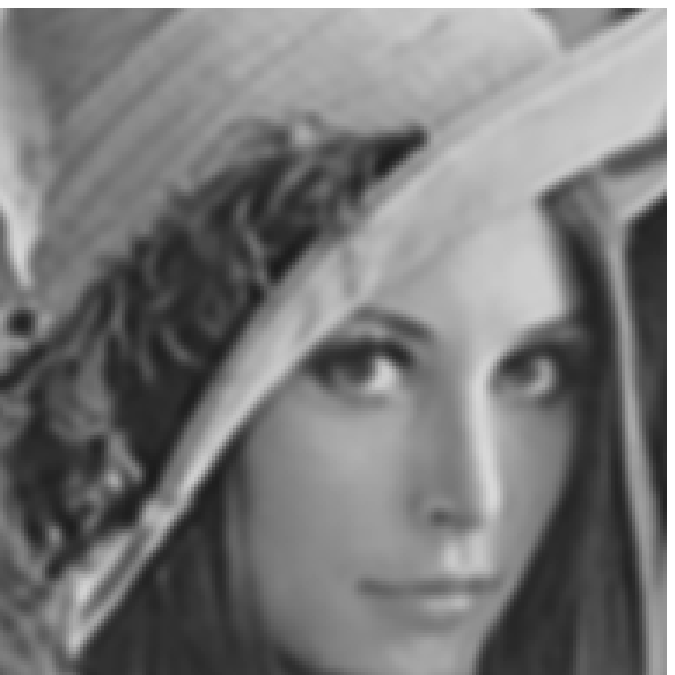}}
{\includegraphics[width=0.2\textwidth]{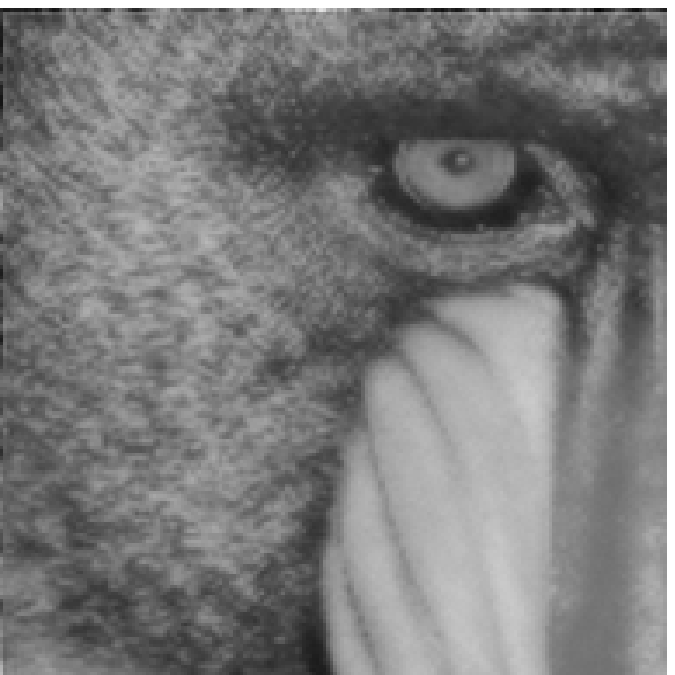}}
}
\mbox{
{\includegraphics[width=0.2\textwidth]{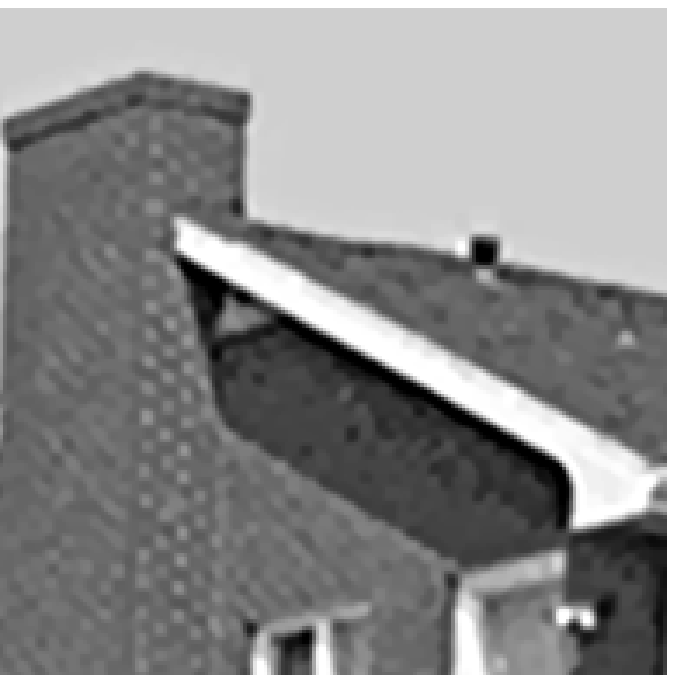}}
{\includegraphics[width=0.2\textwidth]{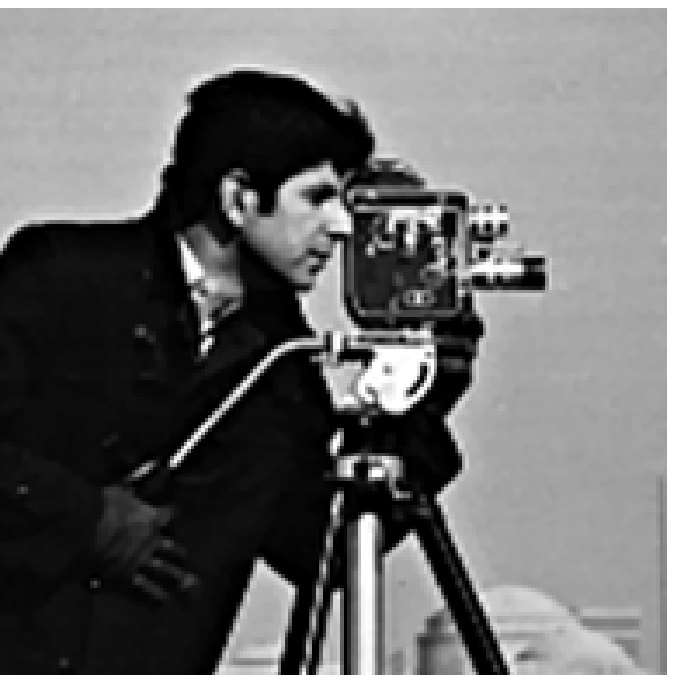}}
{\includegraphics[width=0.2\textwidth]{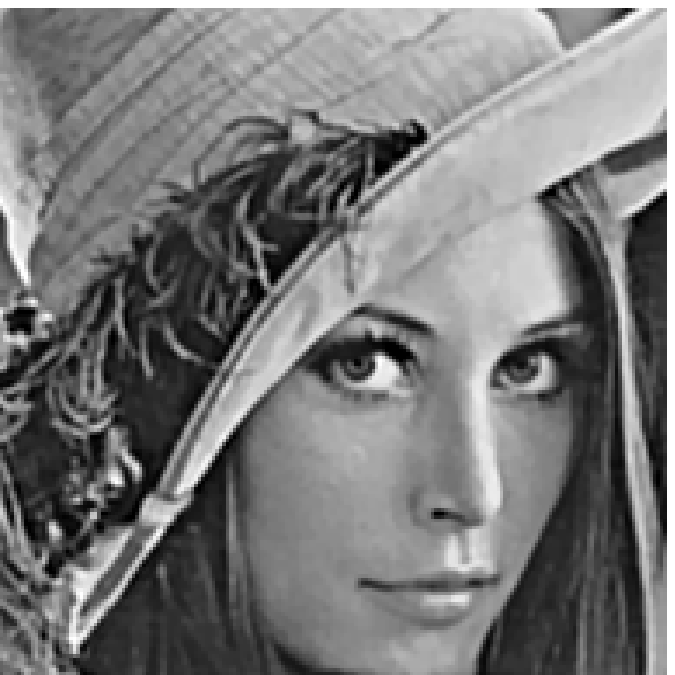}}
{\includegraphics[width=0.2\textwidth]{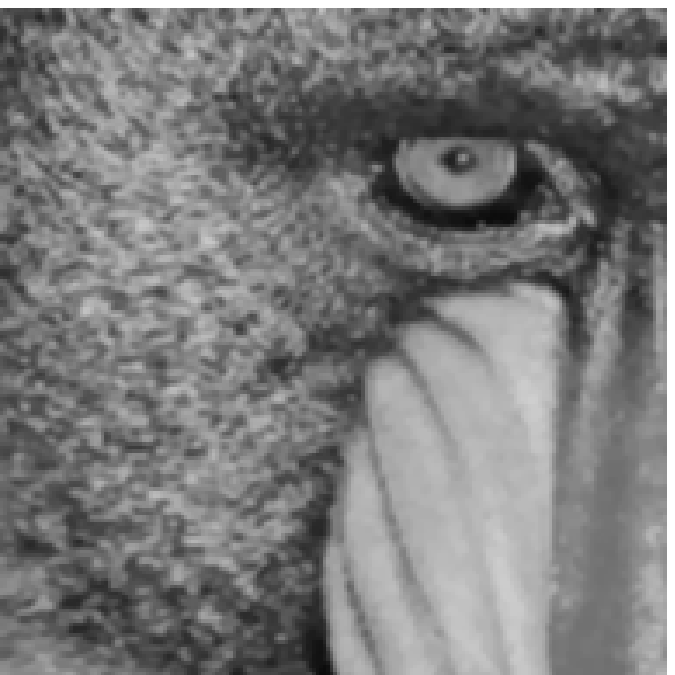}}
}
\mbox{
{\includegraphics[width=0.2\textwidth]{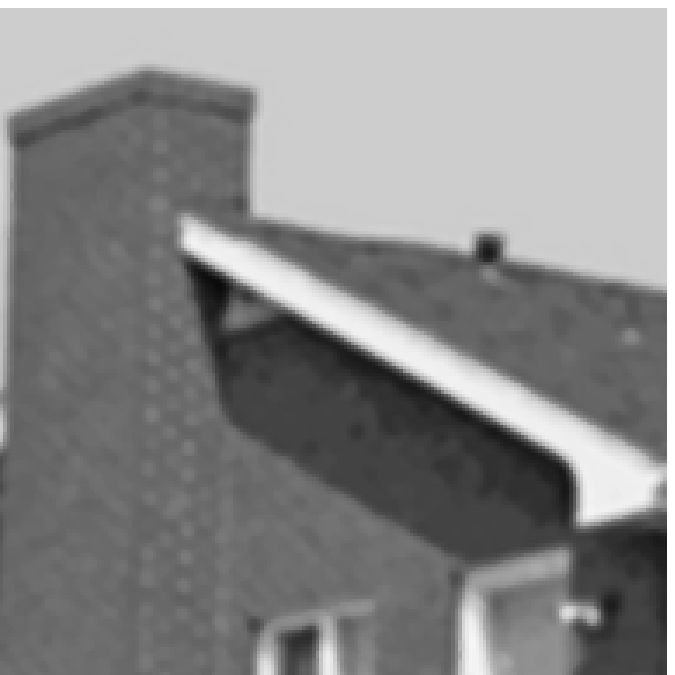}}
{\includegraphics[width=0.2\textwidth]{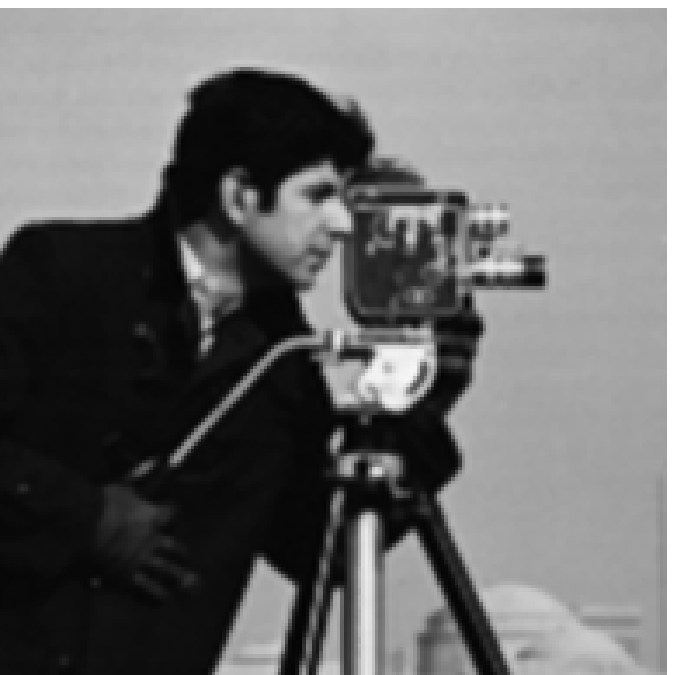}}
{\includegraphics[width=0.2\textwidth]{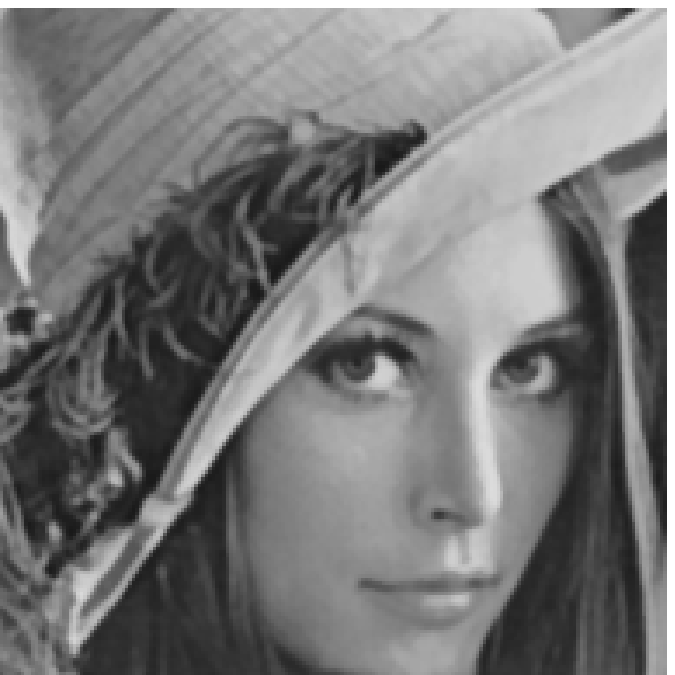}}
{\includegraphics[width=0.2\textwidth]{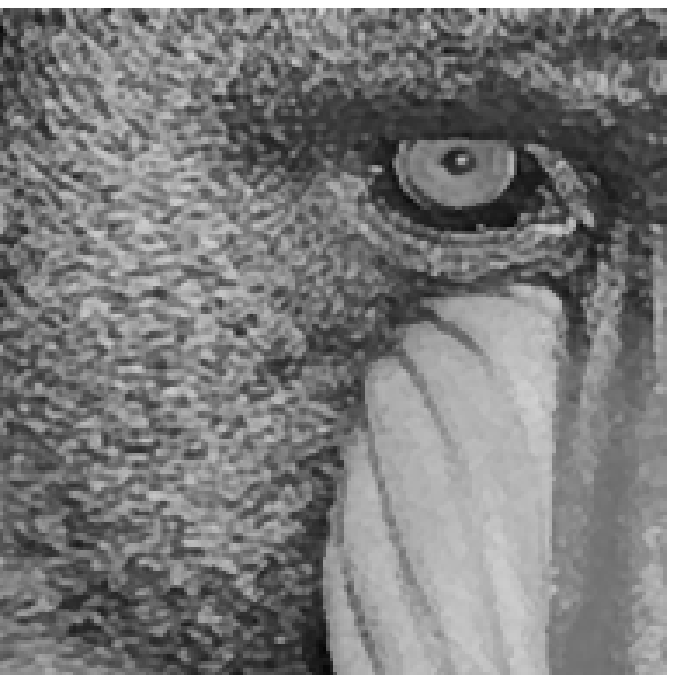}}
}
\caption{Comparison of Deblurring results: (a) $House$ with kernel $1$; (b) $Cameramen$ with kernel $2$;  (c) $Lena$ with kernel $3$; and (d) $Mandrill$ with kernel $4$. From the top row to the bottom one cropped fragments of images are present in the following order: original, blurred, reconstructed by $deconvblind$, Krishnan $et$ $al.$ and the current proposed method. Note that Krishnan's images (the fourth row) looks good, but they are often too sharp to have a high PSNR value. }\label{fig:deblure result}
\end{center}
\end{figure*}


We performed the first experiment in a noise-less environment. The PSNR of the compared algorithms on all test images are shown in Table \ref{table:comparison}. Out of the $16$ test images, four best results are from $deconvblind$, three from  Krishnan et al., and nine from our method. With respect to the SSD values on the estimated blur kernels, our method yields the best estimation in almost all cases. Figure \ref{fig:deblure result} compares the visual quality of the deblurring results of all the methods.

We performed the second experiment in a noisy environment when a white  Gaussian noise at a signal-to-noise ratio of 30 dB is added to the blurred images. The PSNRs and SSDs of all the compared methods of this experiment are shown in Table \ref{table:NoiseComparison}.  
To understand the robustness, for each kernel and each method, we calculated the average PSNR reduction of the four images (Img01, Img02, Img03, and Img04) by 
\begin{eqnarray}
\frac{1}{4}\sum_{i=1}^4 (\text{PSNR at $30$ dB of image $i$}\nonumber\\
 -\text{PSNR at noiseless of image $i$} ), \label{reduction}
\end{eqnarray}
and the results have been presented in Table \ref{table:PSNRdifference}.  
Note that we have removed the comparison with the Krishnan et al.'s method in the table, because if we 
compared the PSNRs in Tables \ref{table:comparison} and \ref{table:NoiseComparison}, the Krishnan et al.'s method would be $3$-$4$ dB in average lower than our method.  
As shown in Table \ref{table:PSNRdifference}, our method has a smaller PSNR reduction for each kernel than that of the $deconvblind$,  indicating more robustness of our method in the restoration of images in a noisy environment. Table \ref{table:PSNRdifferenceWithDeconvblind} compares the average PSNR differences of our method and the $deconvblind$ for each kernel on all test images in both noise-less and noisy environments. Except for the noise-less and Kernel 2 case, our method outperforms  $deconvblind$. The sparsity demonstrates its robustness in the cases of $30$ dB and Kernels 2 and 3, where our method achieves high PSNR gain over $deconvblind$.

\begin{figure}[t]
\begin{center}
\begin{tabular}{cc}
{\includegraphics[width=0.3\linewidth]{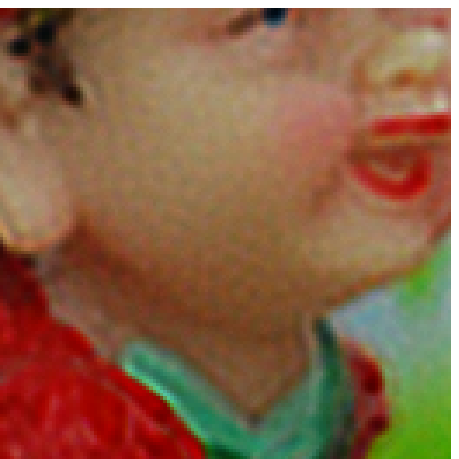}} & 
{\includegraphics[width=0.3\linewidth]{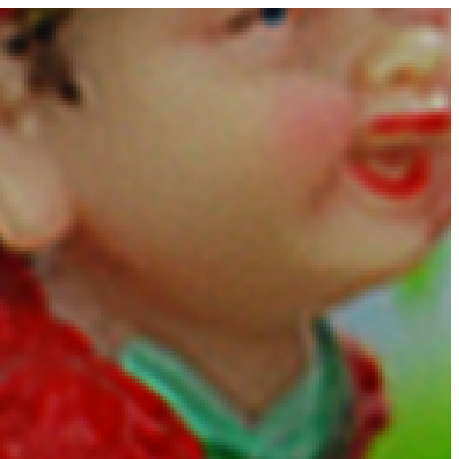}}\\
\end{tabular}
\caption{Highlighted the cheek of bridegroom in Figure \ref{fig:real_life1}. (a) Result of $deconvblind$ and (b) our method. Sparkle artefact can be found in (a).} \label{fig:real_life1_part}
\end{center}
\end{figure}

\begin{figure*}[!ht]
\begin{center}
\begin{tabular}{cc}
{\includegraphics[width=0.35\textwidth]{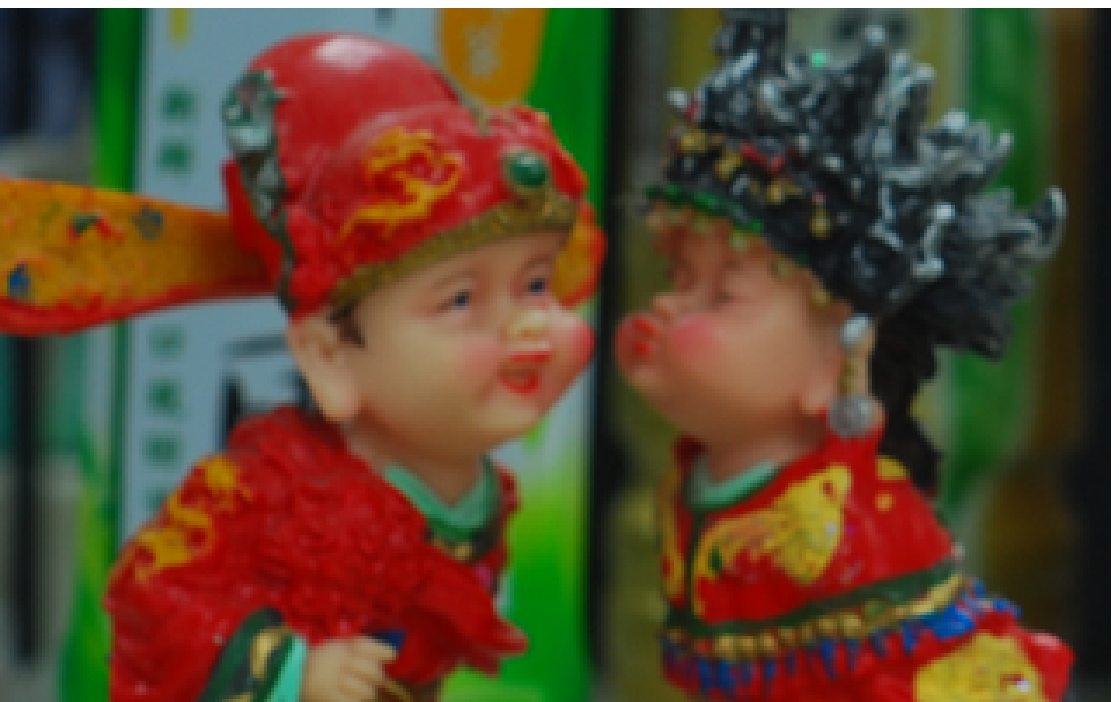}} & {\includegraphics[width=0.35\textwidth]{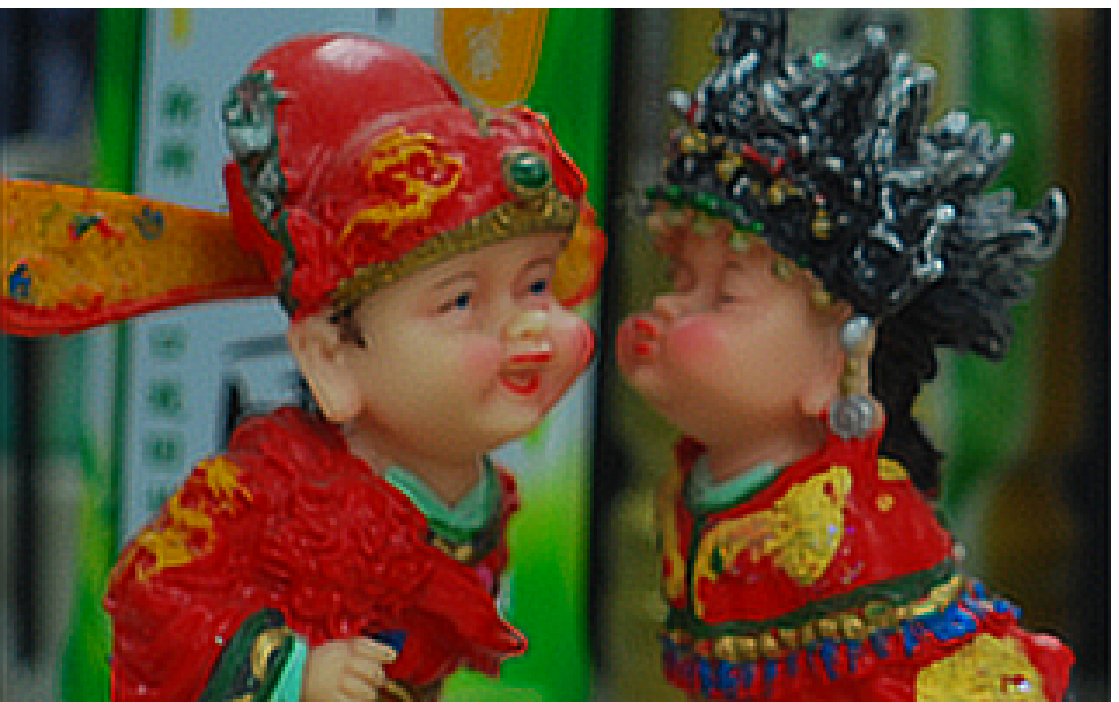}}\\
\end{tabular}
\begin{tabular}{ccc}
{\includegraphics[width=0.35\textwidth]{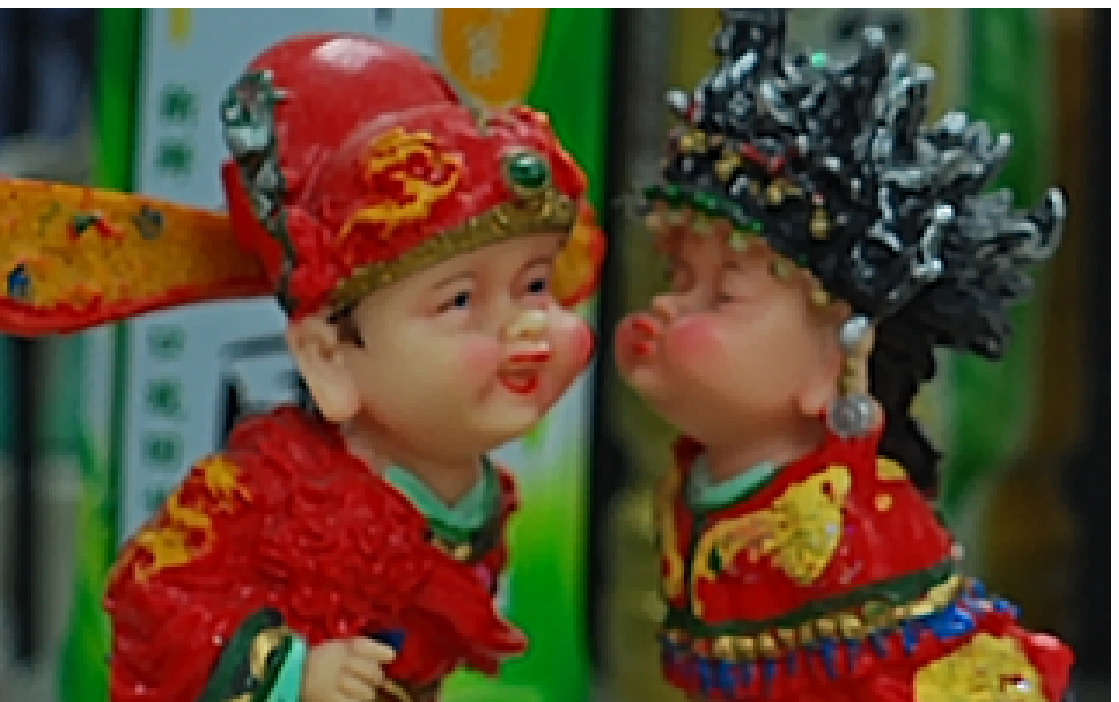}}
& {\includegraphics[width=0.35\textwidth]{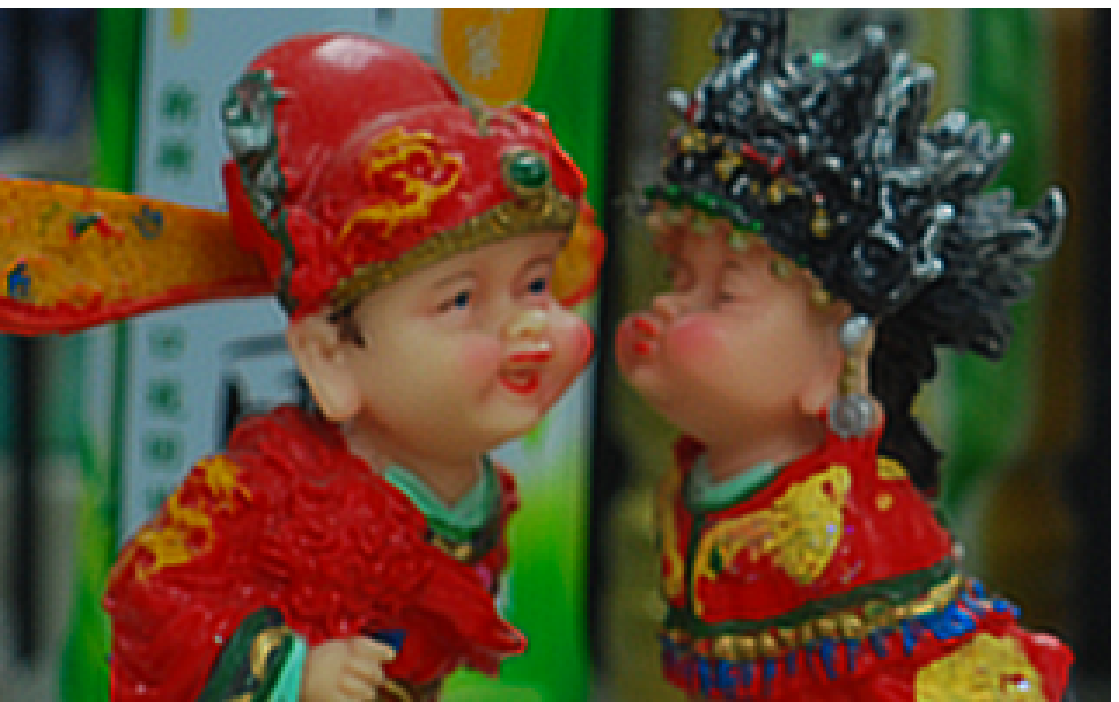}}\\
\end{tabular}
\caption{Results of real-life photographs. (a) Blurred image. (b) $deconvblind$. (c) Krishnan's result. (d) Our result.} \label{fig:real_life1}
\end{center}
\end{figure*}

\begin{figure*}[!ht]
\begin{center}
\begin{tabular}{ccc}
\includegraphics[width=0.24\linewidth]{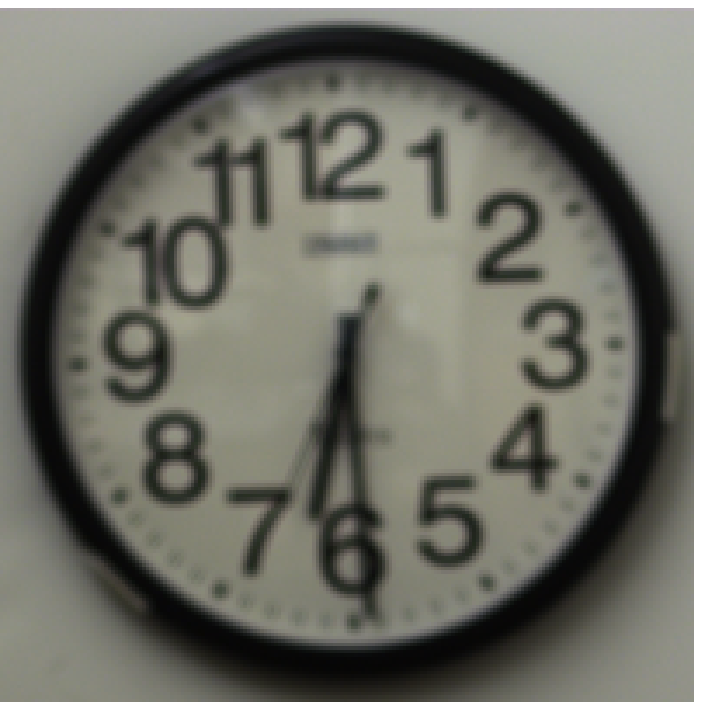}&
\includegraphics[width=0.36\linewidth]{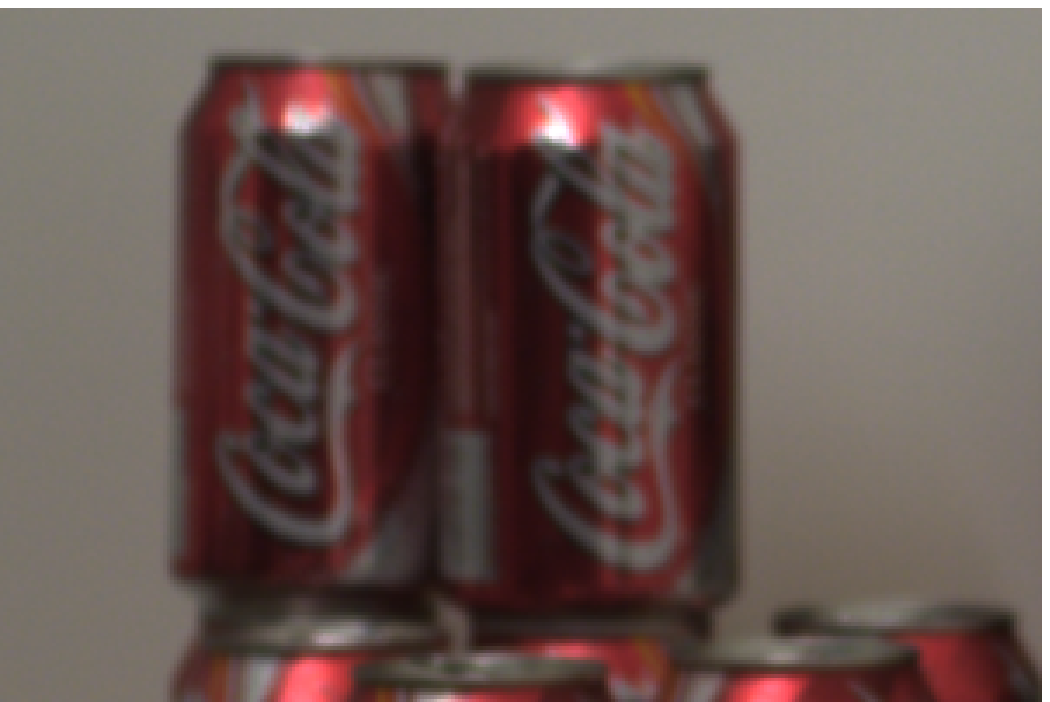}&
\includegraphics[width=0.24\linewidth]{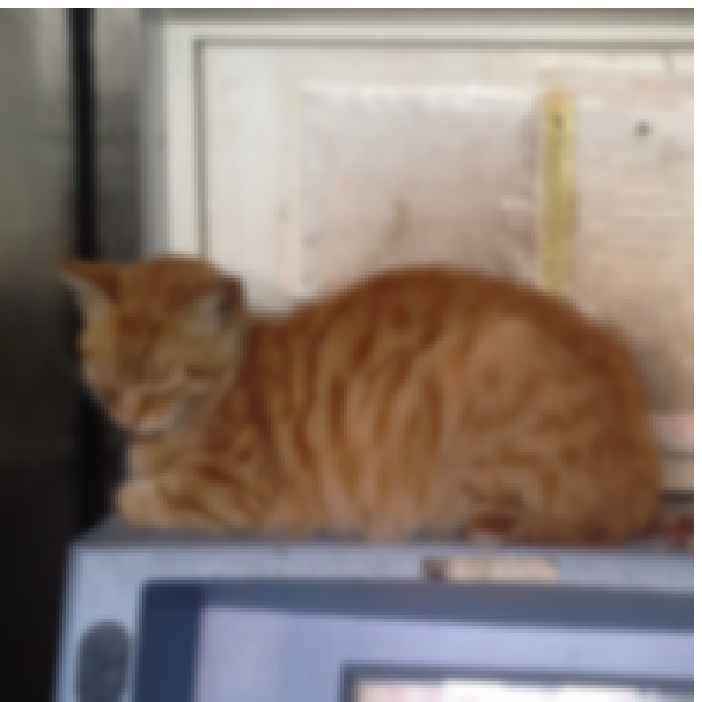}\\
\end{tabular}
\begin{tabular}{ccc}
\includegraphics[width=0.24\linewidth]{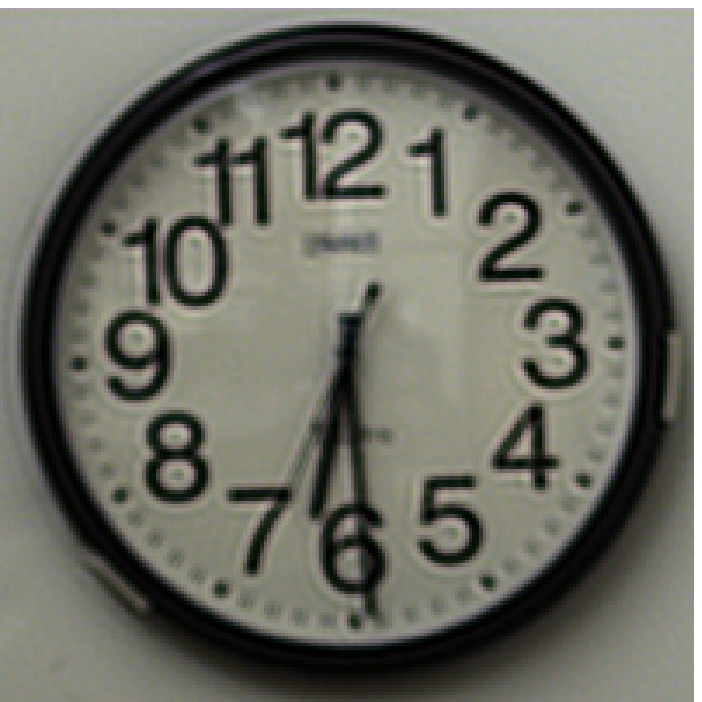}&
\includegraphics[width=0.36\linewidth]{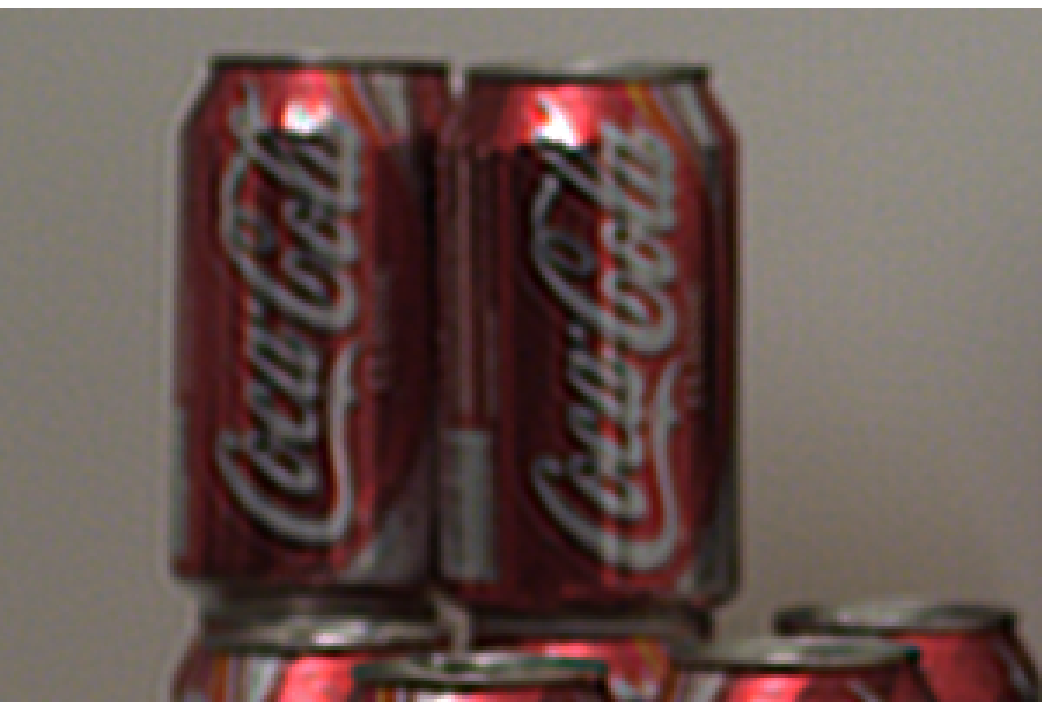}&
\includegraphics[width=0.24\linewidth]{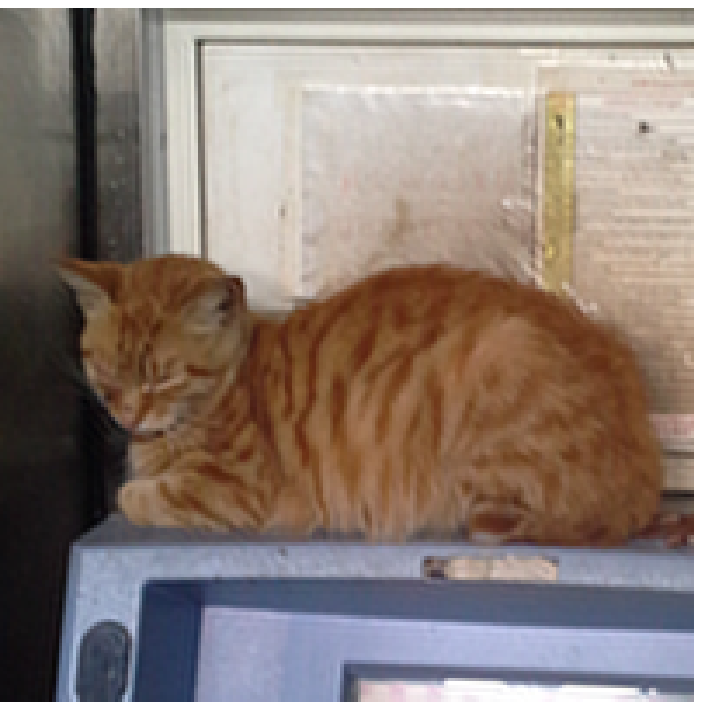}\\
(a) & (b) & (c)\\
\end{tabular}
\end{center}
\caption{Deblurring results of real photographs. Top row: blurred images. Second row: deblurring results of the proposed method.} \label{fig:real_life2}
\end{figure*}

Finally, figures \ref{fig:real_life1} and \ref{fig:real_life2} visually compare the deblurring results of color-images, taken in real-life.
The photographs contain complex structures and different degrees of blurriness. As shown in Figure \ref{fig:real_life1} , the faces of the dolls are clearly restored by all methods.  However,  if we zoomed in the cheek of the bridegroom, as shown in Figure \ref{fig:real_life1_part}, the cheek from our result is smooth while that from the $deconvblind$ has some sparkles in it. Furthermore, figure \ref{fig:real_life2} shows that our method successfully enhanced the sharp edges.

\begin{table}[t]
\begin{center}
\caption{Comparison of the average PSNR reduction for each kernel, according to Equation (\ref{reduction}).}
\begin{tabular}{|c|c|c|c|c|}
\hline
 & Kernel 1 & Kernel 2 & Kernel 3 & Kernel 4\\
\hline 
deconvblind & -2.6850 & -4.6950 & -8.4400 & -5.1300\\
\hline


Our method & -1.6150 & -2.0275 & -1.4550 & -3.6225\\
\hline

\end{tabular}
\label{table:PSNRdifference}
\end{center}
\end{table}

\begin{table}[t]
\begin{center}
\caption{The average PSNR gain of kernels of our method over $deconvblind$ in noise-less and noisy environment. Notice the gains of Kernels 2, 3, and 4 for noisy environment, where Kernels 2 and 3 can be sparsely represented by our dictionary and Kernel 4 is almost sparse.}
\begin{tabular}{|c|c|c|c|c|}
\hline
 & Kernel 1 & Kernel 2 & Kernel 3 & Kernel 4\\
\hline 
noiseless & 0.06 & -0.3150 & 0.95 & 1.6724\\
\hline

30 dB noise & 1.13 & 2.3525 & 7.935 & 3.18\\
\hline

\end{tabular}
\label{table:PSNRdifferenceWithDeconvblind}
\end{center}
\end{table}

\section{Conclusions} \label{sec:con}
Regularization on an unknown blur kernel determines the performance of the blind image restoration problem. In the current paper, we have proposed a novel approach to construct regularization by modelling a blur kernel as a sparse representation of a tensor dictionary, where the dictionary is composed of basic 2-D pattern. Since the dictionary approach has the freedom to be customized for various applications, our approach can be used to connect various regularizations that have been imposed on blur kernels in different applications. 
As a demonstration, we construct a dictionary with atoms formed by the Kronecker product of two 1-D scaled Gaussian functions and show that this dictionary can effectively restore images blurred by the mixed Gaussian types of blur kernels. We also demonstrate that our approach can be efficiently solved by using the variable splitting method for image estimation and proximal gradient method for blur kernel estimation. Furthermore, we compare the performance of our method with some state-of-the-art methods for various sets of images and blur kernels. In most cases, our method derives the best image (in terms of PSNR) as well as blur kernel estimation. An interesting direction for further study is to incorporate a learning procedure to our approach for various applications.

%
%
%

\section*{Acknowledgment}
The authors would like to thank...

\end{document}